\documentclass[acmsmall,screen=true,nonacm ]{acmart}

\AtBeginDocument{%
  \providecommand\BibTeX{{%
    \normalfont B\kern-0.5em{\scshape i\kern-0.25em b}\kern-0.8em\TeX}}}

\setcopyright{acmcopyright}
\copyrightyear{2018}
\acmYear{2018}
\acmDOI{XXXXXXX.XXXXXXX}

\acmJournal{JACM}
\acmVolume{37}
\acmNumber{4}
\acmArticle{111}
\acmMonth{8}
\usepackage{dirtytalk} 
\usepackage{url}

\newcommand{\hide}[1]{}
\usepackage{tikz}  

\usepackage{graphics} 

\begin{document}

\title{Reproducibility of Machine Learning: Terminology,  Recommendations and Open Issues}




\author{Riccardo Albertoni}
\affiliation{%
 \institution{Istituto di Matematica Applicata e Tecnologie Informatiche "Enrico Magenes", Consiglio Nazionale delle Ricerche (IMATI-CNR) }
 \streetaddress{Via De Marini, 6, 16149 Genova}
  \city{Genova}
  \country{Italy}
}
\author{Sara Colantonio}
\affiliation{%
  \institution{Institute of Information Science and Technologies "A. Faedo" of the National Research Council of Italy (ISTI-CNR)}
  \streetaddress{Via G. Moruzzi, 1, 56124 Pisa}
  \city{Pisa}
  \country{Italy}
}
\author{Piotr Skrzypczy\'{n}ski}
\affiliation{%
  \institution{Institute of Robotics and Machine Intelligence, Pozna{\'n} University of Technology}
  \streetaddress{ ul. Piotrowo 3A, 60-965}
  \city{Pozna{\'n}}
  \country{Poland}
}
\author{Jerzy Stefanowski}
\affiliation{%
  \institution{Institute of Computing Science, Pozna{\'n} University of Technology}
  \streetaddress{ ul. Piotrowo 2, 60-965}
  \city{Pozna{\'n}}
  \country{Poland}
}







\begin{abstract}

Reproducibility is one of the core dimensions that concur to deliver Trustworthy Artificial Intelligence. Broadly speaking, reproducibility can be defined as the possibility to reproduce the same or a similar experiment or method, thereby obtaining the same or similar results as the original scientists. It is an essential ingredient of the scientific method and crucial for gaining trust in relevant claims. A reproducibility crisis has been recently acknowledged by scientists and this seems to affect even more Artificial Intelligence and Machine Learning, due to the complexity of the models at the core of their recent successes. Notwithstanding the recent debate on Artificial Intelligence reproducibility, its practical implementation is still insufficient, also because many technical issues are overlooked. In this survey, we critically review the current literature on the topic and highlight the open issues. Our contribution is three-fold. We propose a concise terminological review of the terms coming into play. We collect and systematize existing recommendations for achieving reproducibility, putting forth the means to comply with them. We identify key elements often overlooked in modern Machine Learning and provide novel recommendations for them. We further specialize these for two critical application domains, namely the biomedical and physical artificial intelligence fields.

\end{abstract}

\ccsdesc[500]{Computing methodologies~Machine learning}
\ccsdesc[300]{Computing methodologies~Artificial intelligence}
\ccsdesc[300]{General and reference~Experimentation}
\ccsdesc[300]{General and reference~Evaluation}

\keywords{reproducibility, terminology, recommendations, deep learning, physical artificial intelligence, biomedical applications}

\maketitle
\section{Introduction}


In recent years, Artificial Intelligence (AI) has made rapid progress in terms of new methods, tools and application range. Despite the positive benefits of this development of AI, just in consideration of its impact on individuals' daily life and societies at large, there is also a growing awareness of AI limitations, risks and new challenges that were not so apparent in the previous period. Some concerns also cover the limited people’s trust in the current AI systems applied in real life \cite{Kaur2022}. To increase the positive consequences of AI and, at the same time, mitigate its risks or dangers, several communities are promoting the ideas of Responsible AI or Trustworthy AI, see e.g. their reviews in \cite{BoLi2022,Chatila}. Following these postulates, besides the traditionally considered performance criteria (such as accuracy), many other aspects of AI systems should be taken into account to improve their trustworthiness, see e.g. EU proposal in \cite{EUGuidelines}. The reproducibility of AI methods is among their core requirements and a cornerstone of Trustworthy AI.

Reproducibility in science means that one can repeat or replicate the same (or sufficiently similar) experiment and obtain the same (or sufficiently similar) research results as the original scientists based on their publications and provided documentations. Diverse reproducibility settings and their definitions  have been identified in the literature,  see e.g. \cite{ACMv1.1,Barba,goodman_et_al_2016,Gundersen_2021}, but from a more general standpoint, reproducibility entails that studies are reproduced by independent researchers. Moreover, it is postulated to share sufficient documentation, data and original code, or ability to its proper re-implementation (in the case of a broader meaning of the term replicability).

%

Reproducibility is important for many reasons. Firstly, it is an essential ingredient of science, meant to verify the published results and claims and to enable a continuous self-correcting process. This is crucial for gaining trust in the presented research, but it also allows the community to convert the verified approaches into practice or to build foundations for conducting follow-up new research. In the case of computer science and artificial intelligence, researchers and R\&D engineers should understand new and often very complex methods \cite{Hancox-Li2020}.   They need to check their correctness, examine their working conditions and limitations, and verify the presented results, especially if they want to use them in the applied systems. Then, many AI projects receive either public or business funds, so they should be subject to accountability, and it is necessary to convince others that these projects can produce reliable results, see e.g. discussion in \cite{raff2019step}. Finally, as many AI methods and algorithms are used in critical systems or applications, where their decisions can have an impact on people and society, and their improper operation may cause harm, reproducibility of these methods is one of solutions to test their quality and a signal of their credibility \cite{european_commission_directorate_general_for_communications_networks_content_and_technology_assessment_2020}.

Despite the aforementioned arguments, there is still a lack of details and sufficient reporting in many published works \cite{raw59}. On the one hand, the practice of sharing source code and the postulates of open science are still too slowly developing \cite{stodden_2011,LamprechtGKMAPA20}. However, other reasons are also discussed, such as increasing pressure on scientists to produce and publish results very fast, not having enough time to test the algorithms under different conditions and hyper-parameters, still page limits of some journals or conferences and not encouraging authors to submit supplementary materials or electronic appendixes, too strong bias for presenting only positive results which leads to presenting only selected  details of experiments, reluctance to report failed results, sometimes using questionable research procedures (e.g. in designing experiments, data collection and inappropriate validation), insufficient statistical analysis and testing \cite{Begley,Cockburn,raw59}.



Let us emphasise that concerns about reproducibility come from the growing observations that many of the modern results in science, including AI, in particular in Machine Learning (ML) and Deep Learning (DL) are very difficult to be reproduced. It is even called a reproducibility or repeatability crisis \cite{Cockburn,Heaven2020}. It also occurs in many fields, see e.g. \cite{nosek2022replicability,milkowski2018replicability,Stupple} but notably it is stronger in AI and, in particular ML \cite{Hutson,AI5pilars}.
 This is also supported by analyses of reproducibility elements in published papers from the major AI and ML conferences, which showed that only several dozen percent of them are reproducible; for more details see e.g. \cite{gundersen_kjensmo_2018,raff2019step}. 
 

In our paper, the aspects of reproducibility of research in artificial intelligence will be considered mainly in relation to machine learning. Our focus is due to two reasons. First of all, nowadays machine learning, and deep neural networks in particular, has become one of the most important sub-fields of artificial intelligence, both due to recent research achievements and the success of many applications. Secondly, they include rather complex methods that place much greater demands on reproducibility than methods previously considered in other areas of science. Moreover, ensuring the reproducibility of machine learning is particularly important from the perspective of increasing human trust in the practical solutions of artificial intelligence, which we previously discussed.

Indeed, most of the modern ML approaches, in particular those based on deep neural networks, use very complex models, which are difficult for potential inspection and interpretation by humans. They exploit many hyper-parameters that need specialised and sophisticated optimization strategies. Moreover their training, tuning and evaluation often require additional software libraries accompanied with advanced computational resources, e.g. GPU cards, which may also introduce some nondeterminism and randomness.  In this context, some  authors  postulate that guidelines for the reproducibility of ML and DL complex methods should be extended and provide more technically details than earlier proposed ones or known from other fields of science, see e.g.  \cite{hartley2020,Hancox-Li2020,pineau_et_al_2020,Tatman2018}. 



We see an increase in the interest of reproducibiulity in the AI community, which is reflected in the growing number of papers and the introduction of recommendations for authors' submissions to well-known conferences, such as NeurIPS, ICML or IJCAI, or journals. Nevertheless, in many cases, the perspectives of these best practices are either too general touching upon other issues such as traceability (e.g., \cite{pineau_et_al_2020}) or too specific focusing on data or workflow lineage (e.g., \cite{samuel2020machine}).


Following these motivations, we believe that it is worth writing a survey paper on the reproducibility of artificial intelligence, and machine learning in particular, and their results, addressing researchers interested in these approaches rather than the overly general science perspective.

Firstly, due to the ambiguities in defining the basic concepts of reproducibility, which is visible in the literature, we believe that the terminology issues should be better organized and unified, especially with regard to the type of shared documentation, resources and software environments.
Moreover we claim that it is necessary to compare and harmonize existing best practices and guidelines for reproducibility to provide more specialized information and recommendations based on the above-mentioned difficulties for AI complex models and to extend the ones already identified in the surveyed papers for the context of machine learning and deep networks.

Our survey analyzes many papers on reproducibility issues gathered by means of a systematic literature review with additional elements coming from the narrative review approach. 
We further explain this methodology in the supplementary material to this paper.


The aims of our paper are the following:
\begin{enumerate}
\item To present in section \ref{sec:terminology} a concise terminological review of various definitions of reproducibility, repeatability and replication of research results that we found in surveyed papers. We will attempt to clarify and harmonize terminology and definitions' confusion currently present in the literature. Later in this text we will focus on reproducibility. 
\item To collect and systematically organize the guidelines and recommendations for achieving reproducibility. Unlike many of previous surveys, we want to be more specialized to the perspective of AI/ML researchers. This is why we will approach this issue in two steps. 
\begin{itemize}
\item Firstly, in Section \ref{sec:recommendations}, we will focus on presenting existing recommendations and guidelines from  a more general AI perspective 
by which we plan to identify more precisely the elements  in each of their main categories corresponding to availability of code, data and experiment documentations. In order to show how they could be implemented we will also try to link them to some scenarios or concrete  proposals found in the literature (e.g. metadata sheets or best practices in the literature). Our special contribution will cover three summary tables with them, which should be more comprehensive with respect to AI reproducibility than typically discussed recommendations.
\item Then, in Section \ref{sec:learning}, 
we  will discuss the missing elements and gaps in these general AI recommendations with respect to the characteristics of machine learning reproducibility. In this and the next section \ref{sec:deeplearning}, 
we will discuss the needs for their specialization, especially when considering deep neural networks.
\end{itemize}
\item To discuss in the next two sections \ref{sec:medical} and \ref{sec:physical},
two cases of reproducibility issues in ML-oriented research, concerning namely bio-medical and physical artificial intelligence 
fields, which will additionally show the needs for specializing and extending reproducibility requirements.  We will summarize our proposals of such recommendations  in new summary tables (see section \ref{sec:MLrecommend}).
\end{enumerate}

The aforementioned elements make our survey comprehensive and different from existing surveys. We hope that these elements will contribute to the discussion on reproducibility of machine learning and improve their recommendation both for research and considered applications' perspectives. 
In the rest of the paper,  we have adopted specific text styles to highlight keywords,  the structure of long sections, remarks and open issues to ease the reading. In particular, we have used italics with bold to indicate keywords, bold to partition long sections with subsections' titles making the flow of discussion more evident, and  italics for emphasizing remarks and open issues.

\hide{
\section{Methodology of the review}
\label{sec:method}

This article is written following a \textbf{\textit{systematic survey methodology}}  \cite{siddaway2019,watson2019}, with the aim to synthesise reproducibility guidelines from the existing AI literature into a comprehensive catalogue of recommendations that particularly foster reproducibility practices in modern machine learning.

There are several types of literature reviews that best suit different areas of science or engineering, and support different aims of the conducted review \cite{pare2015}.
A type of review that is focused on revealing  contradictions, inconsistencies, strengths and weaknesses on the given topic in the pre-existing literature is a \textbf{\textit{critical review}} \cite{snyder2019}.

Considering the multi-faceted nature of modern artificial intelligence, 
the choice of the critical review formula appears natural, even if we limit our survey to the area of machine learning.
However, as contemporary AI, and in particular,  machine learning, is characterised by a rapid pace of development, resulting in important findings and recent ideas being reported in non-archival sources, we include in our survey also some elements of a \textbf{\textit{narrative review}}  \cite{greenhalgh2005}. This allows us to summarise what has been written recently about various aspects of reproducibility, thus providing entry points for further individual studies by the readers.
In order to collect the literature for our survey, we have implemented the following steps.
\begin{itemize}
 \item {\bf Search for the literature.}
  We started with a systematic search for the relevant literature in the three most widely used bibliographical data bases:  Elsevier Scopus, Clarivate Web of Science and Google Scholar.
  Recall that Scopus and Web of Science collect mostly archival publications from journals, books and established conferences and have well-defined inclusion criteria for the papers that are covered.
  These databases focus on fundamental and natural sciences, including medicine, but their coverage of engineering disciplines is less comprehensive. This is particularly visible with respect to the recent areas of computer science, where many important conference and workshop publications are not included.
  Therefore we decided to search also the Google Scholar database, which has much more relaxed inclusion criteria.
  Google Scholar indexes recent and emergent sources, which are particularly relevant in the context of the aims of our survey. In addition, it also covers many archiving services collecting the so-called pre-prints or reports, such as e.g. arXiv, the use of which has recently become extremely popular among computer scientists.

 \item {\bf Inclusion criteria.}
  We only included papers written in English that concern machine learning reproducibility.
  Although we included literature from all disciplines, ranging from computer science and  information systems to medicine and natural sciences, we excluded studies on reproducibility for specific topics that did not focused on machine learning or artificial intelligence.
  Only papers published from 2017 to 2022 were included in the search, as we focused on the current state-of-the-art.
  Because each of the databases we considered has a different search mechanism, the queries were slightly different.
  For Scopus we started using the keywords \textit{\textbf{reproducibility AND (machine learning OR artificial intelligence)}}, searching only in the titles and abstracts.
  For each found paper, we initially determined the relevance by the title and abstract, excluding papers that  we deemed irrelevant, e.g. those from medicine which mentioned AI or ML but in fact focused on the  reproducibility of clinical results.
  After this initial pruning of Scopus results, a total of 35 papers were considered for the survey.  For Web of Science, we run the query \textit{\textbf{reproducibility AND artificial intelligence}}, matching the keywords in the title and the abstract of the papers published in the period 2017-2022. We obtained an initial set of 252 papers, whose list was exported and downloaded from the WoS system. Such a set was then checked by analysing the pertinence and relevance of each paper. Thirty-one papers out of the initial 252 turned out to be significant from this check. The excluded papers were either just brief abstract, tutorials or letters or were out of scope as dealing, for instance, with the application of AI in a specific domain (e.g. health and care) and the reproducibility was considered with respect to the generalizability of the domain results. For Google Scholar the search procedure was different  due to the more limited possibilities of asking complex queries with additional conditions and a different way of presenting the results of the search. We asked independently two queries \textit{\textbf{reproducibility of artificial intelligence}} and \textit{\textbf{reproducibility of machine learning}} and received very long lists of results: e.g. 87 600 answers to the first query. As these lists are presented in the relevance order proposed by Google Scholar, we analysed the first 10 web pages - so approx. 100 proposed papers. As a results we identified 39 potentially interesting paper links. After an expert inspection of their full files, we finally selected 28 papers for the deeper analysis.  
 
  Finally, after combining results from the three databases and excluding duplicated papers, we got a list of 81 relevant publications, among them 69 journal articles, with the rest  being conference papers, book chapters or arXiv pre-prints.

\item {\bf Final selection of papers.}
In order to decide which of these papers should be covered in the study
three of the four authors conducted independent evaluations of the 81 full-text papers,
agreeing to select 18 of them as highly relevant to the reproducibility in machine learning problem.
Any controversies while implementing this selection were discussed and resolved.
All these 18 papers are reviewed in our study, while some of the remaining 63
relevant sources found in the databases are cited as secondary examples,
particularly in the context of various applications of machine learning or artificial intelligence and the emerging reproducibility issues.
\end{itemize}

Besides the systematic search for relevant publications we used also a less formal method to get the papers for our survey, namely through backward and forward searches in the references of the papers we have identified in the first phase. 
Here the selection was less formal, as we were interested in highly influential papers on reproducibility with some relevance to machine learning or at least general artificial intelligence, in particular those papers that defined guidelines and recommendations about reproducibility practices. 
In particular, in this scenario we search more intensively the recent conference or journal guidelines or checkpoint lists (see section \ref{sec:guidelines}). 
Our aim in this phase of research was to investigate how and by whom the guidelines we can find in the current literature were proposed, and to find any recurring patterns, inconsistencies, or important gaps in those guidelines. 
Therefore, in this phase we didn't set a publication year census on the papers, including also older but seminal works.

\begin{figure}[htpb!]
    \centering
    \includegraphics[width=0.8\textwidth]{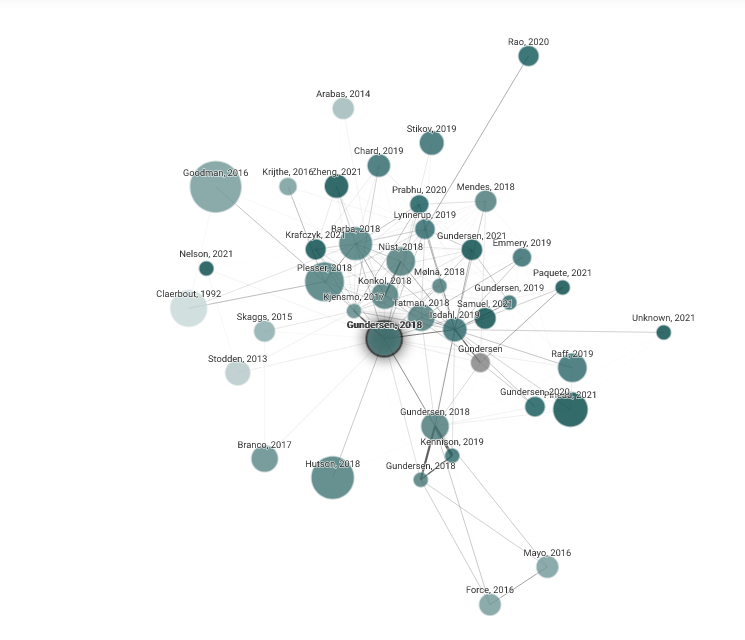}
    \caption{An example graph of connected papers concerning reproducibility,
     which was ``seeded'' by the work of Gundersen and Kjensmo \cite{gundersen_kjensmo_2018}.
     Papers that have highly overlapping references are considered more similar, 
     hence similar papers have strong connecting lines and cluster together. 
     Node size is the number of citations, while node colour is the publishing year (darker nodes are more recent)}
    \label{fig:citegraph}
\end{figure}

The search for most relevant citations in the analysed papers revealed relations between the existing works on reproducibility.
In order to further explore these relations and to identify the most influential papers and authors we explored a new tool to survey the literature and verify what we got.
This web-based visual tool, named Connected Papers \cite{connectedpapers}, allows the users to find papers relevant to a ``seed'' paper, which is entered as the search query.
Related papers are retrieved from the Semantic Scholar Paper Corpus
.
The similarity metric in Connected Papers is based on the concepts of co-citation and bibliographic coupling \cite{coupling2007}. 
Papers that have highly overlapping citations and references are considered as more similar, even if there are no direct citations between them.
The visualised graph clusters the found papers according to their similarity and highlights the shortest path from each node (paper) to the ``seed'' paper in similarity space.
Figure \ref{fig:citegraph} shows an example Connected Papers graph, which was built starting the search from one of the most relevant surveys about reproducibility in AI, the work of Gundersen and Kjensmo \cite{gundersen_kjensmo_2018}. 
In this case 40 papers were used by the program to build the graph, as those having the strongest connections to the origin paper.
When we inspected these papers, we have found that 17 were those already included in our references, while the remaining 23 were less relevant, covering reproducibility or replicability topics in areas other that ML and AI, or not meeting our inclusion criteria.
The graph reveals highly connected clusters of papers around the works co-authored by Gundersen, whom we identify as one of the most influential authors in the survey, and the works of Barba \cite{Barba}, Tatman et al. \cite{Tatman2018}, and Lynnerup et al. \cite{Lynnerup_et_al_2020}. 
Few papers, like the one by Konkol et al. \cite{konkol} are highly connected, but not relevant in our context.
}    
\section{Terminology about reproducibility}
\label{sec:terminology}
Different terms come into play when discussing reproducibility: reproducibility, replicability, and repeatability but also robustness and generability. 
Not all these terms are interchangeable. Authors use the same terms to refer to distinct concepts. Terms result equivalent at a certain level of abstraction but differ when drilling down the steps in the experimental workflow.
To a certain extent, reproducibility terms' proliferation and evolution are evidence of the rich and participated scientific debate. However, \textit{\textbf{redundancy and partial overlaps of terms} slow the communication among practitioners or even distract from the core focus of the reproducibility challenges}. 

This section provides a short overview of reproducibility-related terms to guide the readers in the terminological labyrinth and promote the adoption of more stable and less ambiguous terms. 
We have (i) analyzed an extensive corpus of literature starting and integrating our early review \cite{ReproducibilityTailorHandbookEntry2022}, (ii) identified a set of dimensions to compare the reproducibility-related terms, (iii) clustered the terms according to the dimensions, and  (iv) elaborated some light recommendations for their use.

The considered corpus of literature includes works focusing on scientific reproducibility (e.g., \citet{ACMv1.1,gent_Kotthoff_2014,goodman_et_al_2016,guttinger_2020}), previous terminological reviews (e.g., \citet{plesser_reproducibility_2018, Barba}), and  other works specifically addressing reproducibility in AI, ML and DL (e.g., \citet{Gundersen_2021,Lynnerup_et_al_2020,pineau_et_al_2020}).
Based on the analysis of the literature, we have identified the following dimensions:
\begin{enumerate}
    \item \textbf{\textit{Workflow components availability}} - availability of the components originally deployed in experimental workflows (i.e., data, code and analysis as considered by \citet{pineau_et_al_2020,heil_et_al_2021,gundersen_kjensmo_2018}); 
   \item \textbf{\textit{Teams}} - teams involved in the experimentation (i.e., whether or not the experiments was conducted by the same group that is running the reproducibility validation); 
    \item \textbf{\textit{Reasons}} - reasons because the experiment or part of it is reconducted (i.e.,  validating the repeatability of the experiment or corroborating the scientific hypothesis and theory the experiment aims to support.
\end{enumerate}

\begin{figure}
\begin{tikzpicture}[thick, scale=0.9, every node/.style={scale=0.9}]
\draw (-6,-6) -- ++(6,0) -- ++(0,6) -- ++(-6,0) -- cycle;
\draw (-6,0) -- ++(6,0) -- ++(0,6) -- ++(-6,0) -- cycle;
\draw (0,0) -- ++(6,0) -- ++(0,6) -- ++(-6,0) -- cycle;
\draw (0,0) rectangle (6,-6);
\tiny
\node at (-3,6.5) {SAME DATA};
\node at (3,6.5) {DIFFERENT DATA};
\node[rotate=90] at (-6.5,3) {SAME CODE \& ANALYSIS};
\node[rotate=90] at (-6.5,-3) {DIFFERENT CODE \& ANALYSIS};

\draw (-6,6)rectangle (-3,5);
\node at (-4.5,5.8) {SAME TEAM};
\node[font=\bfseries] at (-4.5,5.5) {Repeatability};
\node[] at (-4.5,5.2) {\cite{JCGM_2012,ACMv1.1,Gundersen_2021, Lynnerup_et_al_2020}};

\node[font=\bfseries] at (-3,4.5) {Reproducibility \cite{pineau_et_al_2020,Claerbout1992,Gundersen_2021, NAS_2019, wieling_et_al_2018,Lynnerup_et_al_2020, Barba}};
\node at (-3,4) {Computational Reproducibility (Silver /Bronze/Gold) \textrm{\cite{heil_et_al_2021}}};
\node at (-3,3.5) {Method Reproducibility \cite{goodman_et_al_2016}};
\node at (-3,3) {Recomputation \cite{gent_Kotthoff_2014}};
\node[gray] at (-3,2.5) {Replicability \cite{stodden_2011}};
\node at (-3,2) {Experiment Reproducible \cite{gundersen_kjensmo_2018}};

\node at (3,4) {Replicability \cite{pineau_et_al_2020, NAS_2019}};


\node at (-3,-2.0) {Independent Reproducibility \cite{raff2019step}};

\node at (-3,-2.5) {Robusteness \cite{wieling_et_al_2018,pineau_et_al_2020}};
\node at (-3,-3) {Data Reproducible \cite{gundersen_kjensmo_2018}};
\draw[fill=white!10](-1.6,-1.6 )rectangle (1.6,1.6);
\node at (0,1.2) {CLOSELY MATCHING} ;
\node at (0,0.9) {ORIGINAL EXPERIMENT} ;
\node at (0,0.4) {\say{Direct} Replications \cite{guttinger_2020}};
\node at (0,-0.1) {Result Reproducibility \cite{goodman_et_al_2016}};


\node[gray] at (3,-2.3) {Reproducibility \cite{islamreproducibility2017, stodden_2011, ACMv1}};
\draw (2.6,-2.6) rectangle (6,-6);
\node at (4.3,-2.8) {CORROBORATION};
\node[fill=white!10 ] at (4.3,-3.3) {Replicability \cite{Peng}};
\node at (4.3,-3.7) {Inferential Reproducibility \cite{goodman_et_al_2016}};
\node[font=\bfseries] at  (4.3,-4.2) {Corroboration \cite{Gundersen_2021}};
\node at (4.3,-4.7) {Conceptual Replication \cite{guttinger_2020}};
\node at (4.3,-5.2) {Method Reproducible \cite{gundersen_kjensmo_2018}};
\node at (4.3,-5.7) {Generalizable \cite{pineau_et_al_2020}};
\node[fill=white!10,font=\bfseries ] at (3,-1.8) {Replicability \cite{ Claerbout1992, ACMv1.1, Lynnerup_et_al_2020}};
\end{tikzpicture}
\caption{Graphical summarization of terms and their main meaning according to the literature and the conceptual dimensions Team, Reasons, Workflow Components. Terms in bold are recommended for use, whereas those in grey are discouraged}
\label{fig:TerminologyDiagram}
\vspace{-0.3cm}
\end{figure}
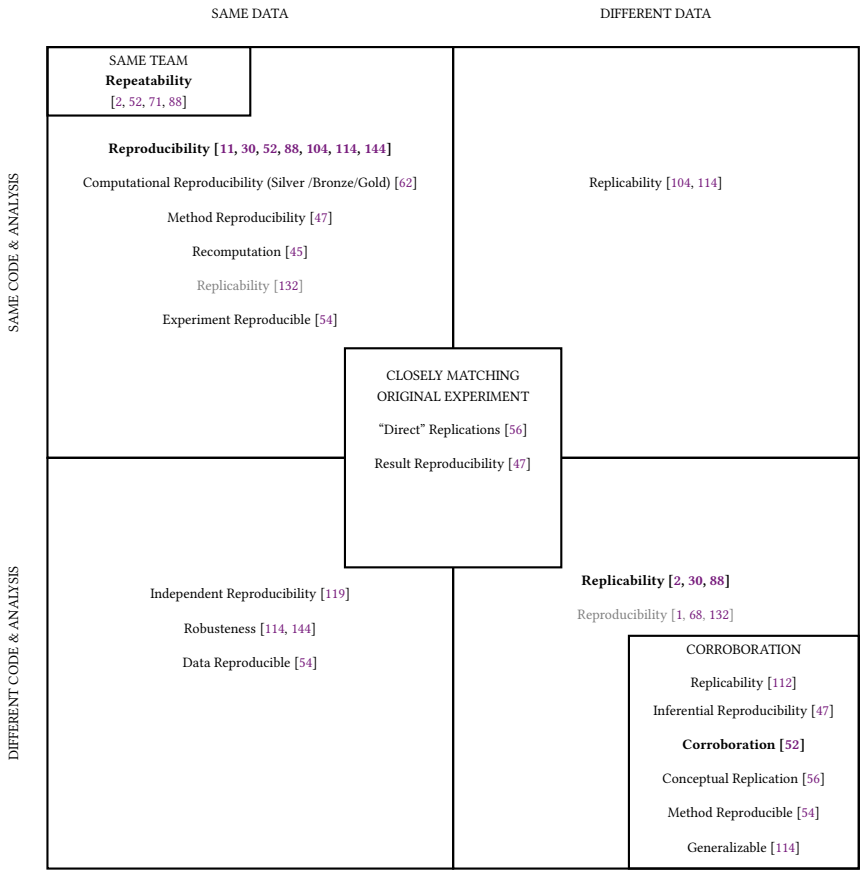

Considering these conceptual dimensions, the reproducibility--related terms used in the literature can be clustered in the following way:
\begin{itemize}
 \item Most of the literature (including \citet{pineau_et_al_2020,Gundersen_2021,NAS_2019,Barba}) refers to reproducibility as the attempt to replicate experiment as much as possible as  the original one, that is by using original data, code and analysis when available. 
    Computational reproducibility, method reproducibility, direct replication and recomputation are used in lieu of reproducibility respectively by \citet{heil_et_al_2021}, \citet{goodman_et_al_2016}, \citet{guttinger_2020}, \citet{gent_Kotthoff_2014}.  
\item The term replicability is highlighted by \citet{Claerbout1992,Lynnerup_et_al_2020,NAS_2019,pineau_et_al_2020}, 
    where an independent team can obtain the same result using the data, which could be slightly different, and methods which they develop completely independently or change slightly.
    Raff \cite{raff2019step} uses in a similar sense the term independently reproducible, describing a situation where the outcome of a paper can be obtained in a statistically similar form by independently implementing an algorithm from this paper.
    Furthermore \citet{Lynnerup_et_al_2020,pineau_et_al_2020} use another name -- robust -- for carrying out the experiments with the same data and some changes in an analysis or code implementations.    
\item Some works such as \citet{ACMv1.1,Lynnerup_et_al_2020,JCGM_2012,Gundersen_2021} uses repeatability to indicate a weaker level of reproducibility where the replication of the experiment is achieved by the same team that provided the original experiments.   
\item  \citet{Gundersen_2021} distinguishes the notion of reproducibility from corroboration. While reproducibility is to produce the result of experiments again or make a copy, corroboration is related to gathering new evidence and make more certain a scientific theory. Some of the analyzed literature adopts terms that indicate some level of corroboration when reconducting an experiment. In particular,  inferential reproducibility, method reproducible, conceptual replication, replicability,  generalizability are used respectively by 
\citet{Peng,goodman_et_al_2016,pineau_et_al_2020,guttinger_2020,gundersen_kjensmo_2018}; 
\end{itemize}
As the first result of our literature analysis, we have delivered the diagram in Figure \ref{fig:TerminologyDiagram}. The diagram summarises terms and their main meaning and the above conceptual dimensions. The diagram is built as a tool to map terms according to their distinguishing characteristics. It might be helpful as a reference for mapping future terminological proposals. It does not represent all the possible combinations of values for the dimensions but highlights as separate boxes those most representative to orienteering into the terminological confusion. The diagram extends the diagram adopted by \citet{pineau_et_al_2020}, mainly concerned with workflow component dimension (e.g., data, code and analysis) with concepts from dimensions that were not considered previously.
In particular, the diagram distinguishes two new boxes for relevant values of the previously unrepresented \textbf\textit{{Teams}} and \textbf{\textit{Reasons}} dimensions: the \say{SAME TEAM} and \say{CORROBORATION}, respectively, to gather terms referring to replications that are conducted by the original team of experimenters and to represent replications aiming at corroborating the experimental hypothesis and theory.
The diagram associates the terms with their supporting references. Similarly to the analysis conducted by \citet{Gundersen_2021}, the terms are not necessarily the same as used in the reviewed resources, and interpretations have been necessary to highlight the gist of the terms' meaning. 


As another result of our literature analysis, we provide some terminological recommendations in Figure \ref{fig:TerminologyDiagram}: we indicate the terms recommended for use in bold. We build upon the existing terms, purposely restraining from minting new terms. Minting new terms would risk adding complexity to an already quite complex picture. In particular, we built upon the concept of reproducibility introduced by \citet{Claerbout1992}, which is a seminal work also adopted in a number of more recent papers  (e.g., \cite{Gundersen_2021,Lynnerup_et_al_2020,pineau_et_al_2020}).
According to this concept, \textbf{\textit{reproducibility}} refers to the ability of an independent researcher to reproduce the same or reasonably similar results using the data and the experimental setup provided by the original authors.
Reproducibility should not be confused with other terms 
such as repeatability and replicability (\cite{plesser_reproducibility_2018}). The term \textbf{\textit{repeatability}} appears in some references, e.g. \citet{ACMv1.1} that uses a notion of reproducibility inconsistent with our definition but should be considered to describe an ability of a researcher to repeat his/her own experimental procedures using same experimental setup and data, while achieving reasonably repeatable results that support the same conclusions.
\textbf{\textit{Replicability}} defined in a way consistent with our understanding of reproducibility is the ability of a typically independent researcher to produce results that are consistent with the conclusions of the original work, using new data or different the experimental setup. 
It is worth clarifying what is reproduced as a result of the above activities and how to understand the term result. 
    Distinguishing among outcome, analysis and interpretation, hypothesis, and theory in an experiment defines the edge between corroboration and other terms in the reproducibility realm. Reproducibility's main aim is to repeat an experiment, while  \textbf{\textit{corroboration}} is more general, it is to support with new evidence or make more certain the conclusion of experiments. Reproducibility is related to experiments, while theories and hypotheses can only be corroborated \cite{Gundersen_2021}.


\section{Reviews of the  recommendations on a general perspective of AI}
\label{sec:recommendations}

This section reviews general recommendations introducing reproducibility levels and their corresponding features to document experiments. 
Firstly, it presents an overview of the related works, including the scientific literature and conference reproducibility guidelines.
Secondly, it deepens this analysis, collecting and harmonizing specific recommendations from a representative literature.
These recommendations will be further extended in the section \ref{sec:learning}, considering issues that arise when considering modern machine learning, related specific technologies and application contexts (i.e., deep learning, cyber-physical systems, and the biomedical images in medical applications).

\subsection{Review of the related works on the reproducibility guidelines}
\label{sec:guidelines}

The texts collected from our systematic review differs in terms of their typology and their recommendation specificity.
We divided them into the following general categories:
\begin{enumerate}
    \item Survey studies or position papers, which also contain some recommendations.
    \item Critical texts analyzing problems of insufficient reproducibility in publications and proposing solutions.
    \item Specialized reproducibility checklists.
    \item Guidelines formulated for paper submissions to major conferences or journals.
\end{enumerate}
\noindent and other texts, such as recommendations for applying machine learning in other, usually applied, fields of sciences. These collected papers are reviewed below following this order. 

Most of the texts discussed so far recognise that ensuring reproducibility for someone’s experiments and results requires proper documentation and sharing of the necessary resources. The documentation should include relevant information which has to be  specified to a certain level of detail. However, there are clear differences of the authors as to the scope and detail of the necessary information. Note that in the case of experimental research in science, it was previously sufficient to describe the procedure, the laboratory environment and the component factors of the experiment to allow this experiment to be repeated by another team. 

Nevertheless, in the case of computer science, and in particular now artificial intelligence, such a paper description is not enough due to the implementation of the method and particular conditions of its running \cite{Peng}. Even in the 90’s, authors such as \cite{Claerbout1992} recommended sharing everything on additional resources (discs in that time) so that anyone could read reports and re-run the provided software and experiments. However, these were not common practices  in the previous century until interest in open science and open source repositories grew. Since then, expectations for the details of documentation have increased. For instance, \cite{goodman_et_al_2016}  recommends to report all relevant aspects of the experimental design, conduct, measurements, data and analysis to achieve the method reproducibility.  Moreover, since the last century, more complex intelligent systems have been developed that definitely require much more experimented effort to fine-tune their parameters than before, especially it concerns modern deep neural networks. 

\noindent 
\textbf{General recommendations from survey studies or position papers}: 
In recent years, the highly influential papers of Gundersen and his co-authors have introduced the division of the available documentation of papers in the field of artificial intelligence into various categories. Following \citet{gundersen_kjensmo_2018} these categories were related to method, data and experiments.

The first \textbf{\textit{method}} category should contain a text description of the AI method, with its high level pseudo-code and Its explanation.  

The \textbf{\textit{data}} category requires sharing the data used in the experiment with an additional text summary. Besides the metadata it should also contain more details on its split into training, testing and validation parts. 

The third \textbf{\textit{experiment}} part -- these authors recommended that a proper experiment text documentation should contain the purpose of the experiment and its setup, the specification of necessary hardware and software used in it, information on independent variables, hyper-parameters of the AI program and other environmental settings. 

Given these general categories, they proposed to distinguish between three \textbf{\textit{different degrees of reproducibility}}:   Experiment Reproducible (R1), Data Reproducible (R2) and Method Reproducible (R3), depending on which documentation was provided to reproduce results.

The R1  degree is the most expected and demanding, where the same implementation of the AI method, data, and the full detailed description of the experiments could be used to produce the same results. It is also called fully reproducible \cite{Peng} or the method reproducibility \cite{goodman_et_al_2016}. R2 degree requires the method description and data (so an alternative implementation should be done), while R3 expects only the method documentation.
 In terms of generality, R1, R2, and R3 stretch from reproducibility to corroboration. R1 reproducible results are less general than R2, which are less general than R3 \cite{gundersen_kjensmo_2018}. Experiment reproducible (R1) corresponds to the reproducibility as defined by \cite{pineau_et_al_2020,Claerbout1992,Gundersen_2021,NAS_2019, wieling_et_al_2018,Lynnerup_et_al_2020} and meant in this survey. Data Reproducible (R2) and Method Reproducible (R3)  aim at testing the generality of results rather than assessing the experiment's reproducibility. 
 In term of documentation required the order is inverted \cite{gundersen_kjensmo_2018}.
 Not having original data, code might be not strictly necessary for generalising an experiment, but it makes it harder to reach Experiment Reproducibility. 
 In general, the increased documentation efforts lead to decreasing efforts to reproduce other research.   Also,  it might be challenging to state anyone has corroborated a theory resulting from a given experiment if he/she cannot access data and codes, and none has ensured the original experiment's reproducibility. 

Gundersen in \cite{gundersen2018aimag} extended these guidelines on different categories by providing more detailed recommendations based on best practices from literature and practical experiences. These were separate lists of detailed recommendations for the following categories: data, source code,  description of the method,  experiments. 
It is further extended in the latest paper \cite{Gundersen_2021}, where the authors extended an interpretation of previous documentations into types of text, code and data. Depending  which of them are shared with independent researchers, these re-defined four types of reproducibility are following: R1 (description), R2 (code), R3 (data), and R4 (experiments which should contain the most complete documentation of the experiments including data, code in addition to the textual description of the experiments with all details and analysis outcome).

The similar categorization of reproducibility could be found in \cite{heil_et_al_2021}'s  proposal, where the three degrees of the \textbf{\textit{reproducibility standards for ML}}, bronze, silver or gold,  are defined basing on availability of data, model, and code, as well as other analyses or programming dependencies.

\noindent \textbf{Critical analyses of insufficient reproducibility in ML publications}: 
Based on the analysis of  ML papers from ICML and NIPS conferences Tatman et al \cite{Tatman2018} distinguished three levels of their reproducibility: low (paper only), medium (sharing code and data) and high (also environment used to run this code over the data) and recommended some practical steps. For instance, for the medium level they advocate of making the project modular, documenting the original environment including all software dependencies, distributing code or data as open source with appropriate licenses, reproducing own project on a new machine, if possible, before sharing it, also they promote using appropriate hosting services for providing the code or data. For high reproducibility they refer to the executable environment with linked all libraries and dependencies necessary to rune the code on a new machine or another operating system. Recall that Peng also referred to improving reproducibility by linking the environment with code and data \cite{Peng}. 

In  \cite{Tatman2018}  recommendations its sharing is discussed in three possible technological options: 
\begin{itemize}
    \item using special hosting services (in particular non-profit ones),
    \item providing a container (in particular dockers, which allow researchers to combine code, data and all environmental dependencies needed to run the code in a single portable formats)  
    \item or a virtual machine image.  
\end{itemize}    
Following their recommendations all these efforts should better support reproducers and try to minimize the number of steps they need to perform. For instance using special scripts that can be called to read data, prepare the appropriate environment and run the code with ideally a single command. Note that  \cite{heil_et_al_2021}'s  proposal of the gold standard also recommends the authors should also prepare this analysis reproducible with a single command - which is certainly the most demanding with respect to full automatization of the reproducibility process.

\noindent
\textbf{Specialized reproducibility checklists}: 
\citet{pineau_et_al_2020} proposed some activities to improve reproducibility in AI and ML community. The best known refers to NeurIPS conference special reproducibility program, including also the submission paper and code procedure and defining the \textbf{\textit{Machine Learning Reproducibility Checklist}} as its part.  In this checklist they specified the necessary elements to be documented and made public with respect to the following categories: model and algorithm, theoretical claims (concerning the rules of writing the paper), datasets used in experiments, shared code including dependencies specifications, all reported experimental results (with all details for the experimental setup, splits of the dataset, hyper-parameters, training details, definitions of evaluation measures, and description of the computing infrastructure used).  

\noindent
\textbf{Guidelines formulated for paper submissions to AI conferences and journals}: 
Similar reproducibility guidelines were also proposed for major AI/ML conferences or scientific journals. The representative examples are IJCAI 2021 Reproducibility Guidelines,   AAAI  conf. Reproducibility checklist (2021) or sub-parts of the reviewer’s instructions for ICML, CVPR. Some of these guidelines are inspired by the above discussed \citet{pineau_et_al_2020} and extend it in some points. In general they require: sharing the code and data as the basic recommendations to authors,  a clear description of model/algorithms - not only its implementation code.  For the data set they also indicate providing its description with respect to  its characteristics and steps of its data cleaning and preprocessing to the final representation. One can notice many differences among these guidelines on how deeply experiments, its parametrization, etc., should be documented. Nearly all documents mention hyper-parameters. Some of them just require to list them, while others ask for more - ranges of their values and how the best ones were tuned. Many documents ask for details of splitting data sets into training, testing and validation ones; precise lists/ definitions of evaluation metrics, number of training and evaluating runs; average training and inference time. 

A few guidelines are even more demanding, e.g, IEEE T-IFS Guides for deep learning submissions \cite{IEEE-TIFS} contains over one page long list with a very detailed checklist of the network topology, types of functions, internal parameters, initialization, random seeds, details of running learning algorithms. It is also more specific to the details computational experiments and the used software.

Note that \cite{Lynnerup_et_al_2020} provide similar recommendations for ML in robotics, by focusing on the reproducibility of computation experiments on real robots.  They also stress the role of managing properly the software dependencies, distinguishing between experimental code and library code, and documenting the measurement metrics, which is essential for reinforcement learning.

\textbf{Recommendations for applying machine learning in other fields of sciences}: 
In some fields of scientific research, other recommendations are also  proposed. For instance, \cite{raw59} contains several postulates for more detailed reporting of results of machine learning when applied in medical and health application (we will discuss them in more details in section \ref{sec:medical}). The position paper \cite{milkowski2018replicability} discusses repeatability, reproducibility and replicability of empirical studies in psychology and computational neuroscience and concludes with some recommendations to improve them - one of their conclusions says that all details relevant to running the code and experiments should be shared even if they seem to be scientifically inessential. Yet another example is  \cite{Artrith}, which provide best practices in machine learning applications in the modern chemistry, where, for example, they pay more attention to the provenance of the data, ensuring their quality and their appropriate pre-processing to the representation relevant for the further analysis.


It should also be noted that most of these guidelines, even in the form of a checklist, are some kinds of recommendations for authors in submitting papers, or indications for editors and reviewers. Despite the inspirations of checklists from other domains, they are rather advices or indications for expected authors’ routing or the organization policy.  However, in some studies some of them are reformulated to more precise \textit{\textbf{requirements}, which are expected to be met}.  

For instance Gundersen et al defined in \cite{gundersen_kjensmo_2018} several \textbf{\emph{metrics}} based on the variables in categories R1, R2 and R3 to quantify reproducibility in their survey of papers coming from the top AI conferences (where each of these paper was evaluated with respect to these metrics).  Another approach toward quantifying reproducibility in machine learning papers was undertaken by Raff in \cite{raff2019step}, where he defined 26 attributes for checking and evaluating each of considered papers.  Recently Gundersen et al. came back to assessing how the available open source and commercial machine learning platform support the reproducibility of experiments \cite{gundersen2022machine}. Similarly to earlier survey \cite{gundersen_kjensmo_2018}, they defined 22 variables for their main categories, which should be verified whether the given platform satisfies them or not, and then three metrics expressing levels of reproducibility were defined on the basis of fulfilling these variables. It allowed them to quantitatively compare and rank these platforms. 

Finally taking about more practical consequences of guidelines, we may also refer to special datasheets by \cite{gebru_datasheets_2020}, which  specify how to document the motivation, composition, collection process, recommended uses for data deployed in the systems and experiments;  model cards by \cite{mitchell_modelcards_2019} ease the description of model's intended use cases limiting their usage in contexts for which they are not well suited; factsheets \cite{arnold_factsheets_2019} provide a template for describing the purpose, performance, safety, security, and provenance information to be completed by AI service providers for examination by consumers.

\subsection{A reasoned summary of recommendations}
\label{sec:reasoned_guidelines}

The section presents a summary of the recommendations starting from the most representative checklists and guidelines for reproducibility we identified in the previous section. We have combined complementary perspectives including reproducibility guidelines from the most prominent AI /ML conferences  (i.e., International Joint  Conference On Artificial Intelligence  (IJCAI 22) \cite{IJCAI22Guidelines} and the official NeurIPS 2020 code submission process by \citet{pineau_et_al_2020}), industry-lead documentation proposals (i.e., datasheets \cite{gebru_datasheets_2020}, model cards \cite{mitchell_modelcards_2019}), and the most up-to-date reproducibility factors adopted by Gundersen et al. \cite{gundersen2022machine} to evaluate the reproducibility platforms.

We organised these extensive set of recommendations along three categories, in the spirit of factors adopted in the levels of reproducibility by \citet{gundersen2022machine, gundersen2018aimag}: Experiment, 
Method, 
and Data.  
For each of these categories, we identified 
\begin{itemize}
\item the reference term (i.e. \textbf{\textit{feature}})
\item the matched recommendation(s)
\item the mean to comply with the recommendation(s), namely via metadata (M), supporting platform (P), or the scientific material, paper, report etc. (S)
\item the source guidelines from which the recommendation comes.
\end{itemize}

We have grouped the alike recommendations from distinct guidelines together. One single recommendation can have more than one supporting source. Hence, an effort was made to balance deduplication of redundant recommendations and the use of original recommendation phrasing.
As the result of our work, the three tables, Experiment (Table \ref{tab:ExperimentRequirements}), Method (Table \ref{tab:MethodRequirements}) and Data  (Table \ref{tab:DataRequirements}), report the recommendations for each of the considered categories.
%
%
%
\begin{table}[h!]
\scriptsize
\begin{tabular}{lp{0.4\linewidth}p{0.1\linewidth}l}
\textbf{Feature} &
  \textbf{Recommendation} &
  \textbf{Where} &
  \textbf{Source guideline} \\
  \hline
 Results &
  Document the results and the analysis. &
  P,S &
  \citet{gundersen2022machine} \\
Hyper-parameters &
  The range of hyper-parameters considered, method to select the best hyper-parameter configuration, and specification of all hyper-parameters used to generate results. For alla the (hyper)parameters used for each model/algorithm and tried development of the paper. &
  S,M &
  \begin{tabular}[c]{@{}l@{}}Pineau’s checklist v2 \cite{pineau_et_al_2020}\\ IJCAI 22 Guidelines \cite{IJCAI22Guidelines}\end{tabular} \\
Training &
  The exact number of training and evaluation runs. &
  S,M &
  Pineau’s checklist v2 \cite{pineau_et_al_2020} \\
Measure &
  A clear definition of the specific measure or statistics used to report results. &
  S,M &
 Pineau’s checklist v2 \cite{pineau_et_al_2020} \\
Central tendency and variation &
  A description of results with central tendency (e.g.mean) \& variation (e.g.errorbars). &
  S,M &
  Pineau’s checklist v2 \cite{pineau_et_al_2020} \\
Run-Time Energy cost &
  The average run time for each result, or estimated energy cost. &
  S &
  Pineau’s checklist v2 \cite{pineau_et_al_2020} \\
Claims &
  State how the analysis supports the claims. &
  P,S &
 \citet{gundersen2022machine} \\
Assumption and restriction &
  All assumptions and restrictions are stated clearly and formally &
  S &
 IJCAI 22 Guidelines \cite{IJCAI22Guidelines} \\
Theoretical tool citation &
  Appropriate citations to theoretical tools used are given &
  S,M &
 IJCAI 22 Guidelines \cite{IJCAI22Guidelines} \\
Formal claims &
  All novel claims are stated formally (e.g., in theorem statements) &
  S &
 \begin{tabular}[c]{@{}l@{}}IJCAI 22 Guidelines \cite{IJCAI22Guidelines}\\ Pineau’s checklist v2 \cite{pineau_et_al_2020}\end{tabular} \\
Proof of claims &
  For any theoretical claim, A complete proof of the claim, Proof sketches or intuitions are given for complex and/or novel results. &
  S &
 \begin{tabular}[c]{@{}l@{}}Pineau’s checklist v2 \cite{pineau_et_al_2020}\\ IJCAI 22 Guidelines \cite{IJCAI22Guidelines}\end{tabular} \\
Justification &
  Justify datasets, method, and metrics &
  P,S &
  \citet{gundersen2022machine} \\
Workflow &
  Workflow representation summarizing experiment execution and configurations. &
  P,M &
  \citet{gundersen2022machine} \\
Workflow execution &
  Workflow execution traces with settings raw, processed, and final data. &
  P,M &
  \citet{gundersen2022machine} \\
Worflow result command &
  For all shared code related to this work, check if you include - README file includes table of results accompanied by precise command to run to produce those results. &
  P &
  Pineau’s checklist v2 \cite{pineau_et_al_2020} \\
Training Code &
  For all shared code related to this work, check if you include -Training code. &
  P &
  Pineau’s checklist v2 \cite{pineau_et_al_2020} \\
Evaluation Code &
  For all shared code related to this work, check if you include -Evaluation code. &
  P &
  Pineau’s checklist v2 \cite{pineau_et_al_2020} \\
Pretraining Code &
  For all shared code related to this work, check if you include -(Pre-)trained model(s). &
  P &
  Pineau’s checklist v2 \cite{pineau_et_al_2020} \\
Hardware &
  Hardware used to conduct the experiment. For all reported experimental results, and especially when run-time critical experiments are carried, check if you include- A description of the computing infrastructure used. &
  P,M &
  \begin{tabular}[c]{@{}l@{}}\citet{gundersen2022machine}\\ Pineau’s checklist v2 \cite{pineau_et_al_2020}\\ IJCAI 22 Guidelines \cite{IJCAI22Guidelines}\end{tabular} \\
Software  and code dependency &
  Document the software and code dependencies &
  P,M &
  \begin{tabular}[c]{@{}l@{}}\citet{gundersen2022machine}\\ Pineau’s checklist v2 \cite{pineau_et_al_2020}\end{tabular} \\
Code repository &
  Shared code in repository. &
  P &
  \begin{tabular}[c]{@{}l@{}}\citet{gundersen2022machine}\\ IJCAI 22 Guidelines \cite{IJCAI22Guidelines}\end{tabular} \\
Code paper appedix &
  All code required for conducting experiments is included in a code appendix &
  S &
  IJCAI 22 Guidelines \cite{IJCAI22Guidelines} \\
Unavailable code &
  Some code required for conducting experiments cannot be made available because of reasons reported in the paper or the appendix &
  S &
  IJCAI 22 Guidelines \cite{IJCAI22Guidelines} \\
Code metadata &
  Include metadata for describing the code. &
  P,M &
  \citet{gundersen2022machine} \\
Code and model license &
  Include a license for code and models &
  P,M &
  \begin{tabular}[c]{@{}l@{}}\citet{gundersen2022machine}\\ ModelCard \cite{mitchell_modelcards_2019}\end{tabular} \\
Citation Export &
  Automatically generate reference for experiment, code, models &
  P,M &
  \begin{tabular}[c]{@{}l@{}}\citet{gundersen2022machine}\\ ModelCard \cite{mitchell_modelcards_2019}\end{tabular} \\
Model Reference &
  Paper or other resources for more information: Where can resources for more information be found? &
  P,M &
  ModelCard \cite{mitchell_modelcards_2019} \\
Model Date &
  When was the model developed? &
  P,M &
  ModelCard \cite{mitchell_modelcards_2019} \\
Model Feedback &
  Feedback on the model: E.g., what is an email address that people may write to for further information? &
  P,M &
  ModelCard \cite{mitchell_modelcards_2019} \\
Code citeable &
  Generate a digital object identifier (DOI) or persistent URL (PURL) for experiment. &
  P,M &
  \citet{gundersen2022machine} \\
\end{tabular}
\caption{Recommendation for experiments }
\label{tab:ExperimentRequirements}
\vspace{-0.6cm}
\end{table}
\begin{table}[th]
\scriptsize
\begin{tabular}{lp{0.4\linewidth}p{0.1\linewidth}l}
\textbf{Feature} &
  \textbf{Recommendation} &
  \textbf{Where} &
  \textbf{Source guideline} \\
  \hline
Hypothesis &
  Document the hypotheses to be assessed. &
  P,S &
  \citet{gundersen2022machine}  \\
Prediction &
  Document the predicted outcome. &
  P,S &
  \citet{gundersen2022machine}  \\
Setup &
  Parameters and conditions to be tested and desired statistical significance of results. &
  P,S &
  \citet{gundersen2022machine}  \\
Problem description &
  Description of problem to be solved. &
  P,S &
  \citet{gundersen2022machine}  \\
Outline &
  Describe method conceptually. &
  P,S &
  \citet{gundersen2022machine}  \\
Model involved &
  What type of model is it? This includes basic model architecture details, such as whether it is a Naive Bayes classifier, a Convolutional Neural Network, etc. This is likely to be particularly relevant for software and model developers, as well as individuals knowledgeable about machine learning, to highlight what kinds of assumptions are encoded in the system. &
  P,S, M &
  ModelCard \cite{mitchell_modelcards_2019} \\
Method Assumption and Restriction &
  For all models and algorithms, A clear explanation of any assumptions &
  S &
  Pineau’s checklist v2 \cite{pineau_et_al_2020} \\
Mathematical Setting &
  For all models and algorithms,  A clear description of the mathematical setting, algorithm, and/or model. &
  S &
  Pineau’s checklist v2 \cite{pineau_et_al_2020} \\
Algorithm Complexity &
  For all models and algorithms, An analysis of the complexity (time, space, sample size) of any algorithm. &
  S &
  Pineau’s checklist v2 \cite{pineau_et_al_2020} \\
Pseudo code &
  Support for pseudo code. &
  P,S &
  \citet{gundersen2022machine}  \\
\end{tabular}
\caption{Recommendation for methods. The first and second columns summarize what to describe; the third is where the description is likely to be provided (i.e. in metadata (M), the platform(P), or the scientific material, paper, report etc. (S)); the fourth column includes the guidelines from which the recommendation comes }
\label{tab:MethodRequirements}
\vspace{-0.6cm}
\end{table}
\begin{table}[th]
\scriptsize
\begin{tabular}{lp{0.3\linewidth}p{0.1\linewidth}l}
\textbf{Feature} &
  \textbf{Recommendation} &
  \textbf{Where} &
  \textbf{Source guideline} \\
  \hline
Data repository &
  Share data in a community repository or the simulation enviroment &
  P &
  \begin{tabular}[c]{@{}l@{}}\citet{gundersen2022machine} \\ Pineau’s checklist v2 \cite{pineau_et_al_2020}\\ IJCAI 22 Guideline \cite{IJCAI22Guidelines}\end{tabular} \\
Data distribution &
  How will the dataset will be distributed (e.g., tarball on website, API, GitHub)? &
  P,M &
  Datasheets \cite{gebru_datasheets_2020} \\
Data appendix &
  All novel datasets introduced in this paper are included in a data appendix &
  S &
  IJCAI 22 Guideline \cite{IJCAI22Guidelines} \\
Dataset from literature &
  All datasets drawn from the existing literature (potentially including authors’ own previously published work) are publicly available &
  P,S,M &
  IJCAI 22 Guideline \cite{IJCAI22Guidelines} \\
Cite Data &
  All datasets drawn from the existing literature (potentially including authors’ own previously published work) are accompanied by appropriate citations &
  P,S,M &
  IJCAI 22 Guideline \cite{IJCAI22Guidelines} \\
Data citeable &
  Generate DOI or PURL. &
  P,M &
  \begin{tabular}[c]{@{}l@{}}\citet{gundersen2022machine} \\ Datasheets \cite{gebru_datasheets_2020}\end{tabular} \\
Data relevant statistic &
  For all datasets used, The relevant statistics, such as number of examples &
  P,S,M &
  Pineau’s checklist v2 \cite{pineau_et_al_2020} \\
Unavailable Dataset Description &
  All datasets that are not publicly available (especially proprietary datasets) are described in detail &
  S &
  IJCAI 22 Guideline \cite{IJCAI22Guidelines} \\
Data collection, annotation  and quality &
  For all datasets used, For new data collected, a complete description of the data collection process, such as instructions to annotators and methods for quality control &
  P,S,M &
  \begin{tabular}[c]{@{}l@{}}Pineau’s checklist v2 \cite{pineau_et_al_2020}\\ Datasheets \cite{gebru_datasheets_2020}\end{tabular} \\
Train/validation/test splits. &
  For all datasets used, The details of train/validation/test splits. &
  P,M &
  Pineau’s checklist v2 \cite{pineau_et_al_2020} \\
Excluded data &
  For all datasets used, An explanation of any data that were excluded, and all pre-processing steps. &
  P,S,M &
  Pineau’s checklist v2 \cite{pineau_et_al_2020} \\
Preprocessing cleaning and labelling &
  Was any preprocessing/cleaning/labeling of the data done (e.g., discretization or bucketing, tokenization, part-of-speech tagging, SIFT feature extraction, removal of instances, processing of missing values)? If so, please provide a description. &
  P,S,M &
  Datasheets \cite{gebru_datasheets_2020} \\
Raw data &
  Was the “raw” data saved in addition to the preprocessed/cleaned/labeled data (e.g., to support unanticipated future uses)? If so, please provide a link or other access point to the “raw” data. &
  P,S,M &
  Datasheets \cite{gebru_datasheets_2020} \\
Preprocessing software &
  Is the software used to preprocess/clean/label the instances available? If so, please provide a link or other access point. &
  P,S,M &
  Datasheets \cite{gebru_datasheets_2020} \\
Data metadata &
  Include basic metadata describing the data. &
  P,M &
  \citet{gundersen2022machine}  \\
Dataset contacts &
  How can the owner/curator/manager of the dataset be contacted (e.g., email address)? &
  P,M &
  Datasheets \cite{gebru_datasheets_2020} \\
Data license, Intelectual property, term of use &
  Give the data a license including Intelectual property and use terms or regulatory restrictions &
  P,M &
  \begin{tabular}[c]{@{}l@{}}\citet{gundersen2022machine} \\ Datasheets \cite{gebru_datasheets_2020}\end{tabular} \\
\end{tabular}
\caption{Recommendation for data. The first and second columns summarize what to describe; the third is where the description is likely to be provided (i.e. in metadata (M), the platform (P), or the scientific material, paper, report etc. (S)); the fourth column includes the guidelines from which the recommendation comes }
\label{tab:DataRequirements}
\vspace{-0.4cm}
\end{table}
Overall, the tables collect and harmonize an extensive set of recommendations providing a more comprehensive view than those typically discussed in every separate checklist. Some aspects in the list of recommendations might deserve more attention than others, depending on the nature of the experiments and the kind of technology involved. 

Moreover, not all the recommendations are equally straightforward to be implemented. 
The \textbf{\textit{scientific material}}, paper and report (S) includes typical documentation practice in the scientific community. 
Specific guidance is usually provided by conferences and journal guidelines and very often are requirement for the acceptance. Researchers are at least in their general aspect rather accustomed to them. 
Hence, they might be easier to take up and we will not discuss them in more detail. 
 Whereas recommendations to be implemented via a  platform (P) or metadata (M) require specific supporting actions such as adopting third-party platforms and metadata models, which might deserve a more specific discussion.
For this reason, we briefly discuss supporting platforms and metadata models.

\textbf{Supporting platforms.} As for to what extent existing platforms support these recommendations,  \citet{gundersen2022machine} surveyed  13 machine learning platforms based on its subset of recommendations.  \citet{gundersen2022machine} has chosen the platforms based on the literature on reproducibility including  OpenML, MLflow, Kaggle, Codalab, StudioML, plus the most commonly used machine learning platforms provided by Amazon, Microsoft and Google. 
Though Gundersen et al.'s recommendations are a courser subset of those included in the tables presented in this paper, the lack of support exhibited by Gundersen's analysis might provide some insight to spot open issues.
In particular,  according to  \cite{gundersen2022machine}, none of the considered platforms supports the statement of how the analysis supports the claims. Among the less supported features  \citet{gundersen2022machine} includes the code, data, experiment citation;  the native provision of code and data repositories;  the provision of metadata.    
According to \cite{gundersen2022machine}, platforms do not necessarily need to include all the features natively. They might rely on third-parties services to integrate these functionalities. For example, \textbf{\textit{source code management systems}} such as Git and GitHub can be adopted for sharing data and code. \textbf{\textit{Permanent URI}} to make data sets and code citeable can be mint via services like Zenodo, 
Figshare, 
W3ID 
and Datacite. 
However, \textit{\textbf{combining functionalities provided by different parties} is often left to the experience of those who share the experiment. It might involve a set of different practices, which, if not standardized, might be disorienting for reproducers that are not necessarily into data stewardship and software engineering}. 
 
\textbf{Metadata.} As for metadata, \citet{Gundersen_2021} suggests the provision of basic metadata only, but the tables show that many other features can be tracked via metadata. \say{The FAIR Principles have pointed out that metadata is a crucial aspect when making an experiment reproducible. Over the past few years, awareness of this challenge has led to efforts to establish standards and procedures for better data and metadata collection and sharing practices} \cite{Bittner_2021}.   In particular, a quite rich set of standards have emerged in the context of open data Government and open science. For example, the World Wide Web Consortium (W3C) has developed \textbf{\textit{machine-actionable metadata models}} for documenting provenance (PROV-O \cite{PROV-O}), datasets (DCAT \cite{vocab-dcat-2}),  quality annotations and metrics for data and other kinds of resources (DQV \cite{VOCAB-DQV, AlbertoniDQV}).     \citet{DBLP:journals/fgcs/GarijoGC17} have  provided a holistic, Linked Data compliant, and ready-to-use solution to document \textbf{\textit{workflow specifications and their executions}}, which exploits PROV-O \cite{PROV-O}, P-PLAN (\cite{upm19478}) and the Open Provenance Model for Workflows (OPMW).
\citet{mora-cantallops_traceability_2021} have discussed the importance of metadata models for traceability in AI.

Although not specific to AI experiments and systems, the models mentioned above offer some excellent standing and a backbone for describing data, actors, other kinds of entities, and how these might relate in experiments. 
Such a standing needs to be refined and extended to capture the gist of specific AI experiments. AI-related controlled terminologies are required, for example, to complement the backbones with the hyper-parameters, tasks and metrics for AI techniques. Adopting a backbone, which is defined according to linked data best practices, offers the ability to combine different models and terminologies as needed, easing the tailoring of the such backbone with the required AI-specific and community-governed refinements. Some AI-specific and community-governed metadata models have been proposed (see  ML-Schema \cite{Publio_2018}, Amazon ML Artefact description \cite{Schelter2017}, MEX \cite{Esteves_15}). However, \textit{\textbf{shared AI-related metadata standards} have not yet been established. AI-specific and community-governed metadata models hardly relate to more general open science metadata standards, often reinventing patterns for dealing with trans-domain entities. The current proposals are far from being broadly adopted when documenting AI experiments}.
The lack of standardized specialization for describing the artefacts involved in the AI experiments and their descriptions makes moving experiments among platforms difficult. In contrast, ensuring the portability of experiments and technology might be crucial to satisfy specific constraints arising from scientists and developers and non-academic applications with particular constraints and preferences on the technology to use.

\section{Machine learning -- specialising the requirements}
\label{sec:learning}
 
Nowadays, machine learning, and deep learning in particular is considered the most prominent area in artificial intelligence, due to its wide applications in solving practical problems in a number of areas, from recommendation systems, through medical diagnostics to industrial systems and autonomous agents. Therefore, ensuring reproducibility in ML research becomes particularly important and urgent task, as irreproducible papers and their research artifacts (e.g. neural network models) proliferated to the level of production systems hamper the efforts towards trustworthy AI \cite{williams2022}, making it harder to ensure robustness, traceability and accountability in real-world applications\footnote{OECD AI Principles: \url{https://oecd.ai/en/dashboards/ai-principles/P8}} , and in particular safety-critical domains where human welfare may be harmed (e.g. medical diagnostics, autonomous driving) \cite{grundy2022}.
The following paragraphs provide the characteristics of ML and DL in the context of reproducibility,  consider the features of publications that might be positively correlated with better reproducibility, and extract from the literature the issues that can render irreproducible even a well-documented ML research.  

\textbf{ML \& DL characteristics.} Althought machine learning covers different categories of methods, the reproducibility issues largely relate to the currently most popular approaches to creating complex systems. In addition to neural networks, they also include ensembles of predictive models or hybrid solutions integrating multiple algorithms. Recall that they are often considered the so-called \textit{\textbf{black boxes}} in not offering users information on how they operate inside, what they learned and how they came to a specific decision \cite{guidotti2018survey}. 
The \textit{\textbf{lack of interpretability} of models, not supporting of understanding of their more and more sophisticated operations, combined with the processing of increasingly larger and complex data, additionally reduces the reproducibility of such machine learning methods.} 

Many of the popular ML approaches, in particular those using deep neural networks, are not following the model-driven paradigm. 
They rather result from intensive, repeated experiments with data, in which the model architecture is constructed, along with quite complex parameter tuning. 
Moreover, modern ML methods often require the use of special programming languages, optimisation libraries and hardware solutions such as Graphical Processing Units (GPUs) to accelerate computations. 
This experiment-and-data driven approach also requires wider reproducibility recommendations than those considered in other areas of artificial intelligence.
Therefore, this section is concerned with reproducibility in modern machine learning with neural networks, as this shows especially well the need to extend the requirements discussed in the previous section.

The literature we have collected for this review unanimously agrees that the level of reproducibility in current machine learning research is not satisfactory \cite{gundersen2018aimag,gundersen2019aimag,pineau_et_al_2020,lopresti2021,grundy2022}.
Tatman {\em et al.} \cite{Tatman2018} were among the first, who considered irreproducibility an issue for practitioners applying a new ML method within their own domain.
\textit{They point out that non-academic ML practitioners often consider reproducing the
results of a paper the first step in applying the method described there to new data or an entirely new domain}.
Hence, being able to reproduce these results to a reasonable degree provides an evidence that the method is worth to be considered, which is also being discussed by other researchers, see e.g. \cite{Hancox-Li2020,pineau_et_al_2020}.

\textbf{What makes a ML/DL paper reproducible?} However, while the short paper of Tatman {\em et al.} gives an important motivation for considering the practical side of reproducibility in machine learning, it lacks a statistical evidence to support the claims, and sets the directions towards improved reproducibility, rather than provides detailed recommendations.
Conversely, the work of Raff \cite{raff2019step} focuses on the statistics from investigating reproducibility of 255 ML-related papers published from 1984 until 2017.
Raff correlates the reproducibility of a paper with some of its features, using 
the notion of being ``independently reproducible'', which means the author was able to independently implement an algorithm from the investigated paper and reproduce the majority of claims in this paper applying reasonable and standard libraries.
While 26 features are defined, some of them are highly subjective, such as paper readability or rigorous versus empirical research type.
Among the more objective features (listed as unambiguous or mild subjective in \cite{raff2019step}) the availability of pseudo code, specification of hyper-parameters and computing resources, and a higher number of tables and equations were determined as those being positively correlated with reproducibility.
Interestingly, this study shows that releasing the code is not sufficient for independent reproducibility.
According to Raff's comment ``the inability to reproduce results without code availability may suggest problems with the paper'' due to an insufficient explanation of the method or the lack of important implementation details.

This  stands in opposition to the earlier survey analysis of reproducibility of 400 research papers in AI main conferences from 2013-2016 by Gundersen and Kjensmo \cite{gundersen_kjensmo_2018} which identified 20 binary variables referring to their categories of documentations. However they only identified if these variables are represented in the texts of these papers, and did not re-implement methods or used codes in experiments. Their results show that only between 20\%-30\% of these variables are sufficiently documented. Here we may repeat \cite{Tatman2018} that \cite{gundersen_kjensmo_2018} did not investigate if their variables actually impact the reproducibility. 
\textit{The more pragmatic view of machine learning, as stated in \cite{Tatman2018} is concerned with the question what artifacts should be included with a publication to allow others to understand the method and to apply it in their own work or research?}

Also Lopresti {\em et al.} \cite{lopresti2021} focus on improving the chances to reproduce a research paper by
providing better documentation and explanation of the used ML pipeline. Their investigation concerns entries in competitions associated with conferences in patter recognition and document analysis.
Authors claim that while reproducibility of the paper's outcome (e.g., predictions) using the original artifacts (code, data) is often achieved during the competitions ran on pattern recognition conferences, most of these competitions do not achieve replicability by an independent team using methods which this team developed on the basis of the papers. 
Therefore, they focus on the documentation of the methods and experiments, acknowledging the conclusion of Raff, that making the source code available does not guarantee reproducibility, because additional meta-parameters and details of the third-party or system-level software might be needed to obtain an outcome of the experiment which is similar enough to the original results. 
\textit{If authors who publish code do not document their experiments well, then sharing the artifacts  per se does not improve the chance of reproducing these experiments}. 

From the critical literature analysis carried out in this paragraph, we conclude that no study published so far provides a definitive answer to the question of what features of a scientific paper are those highly correlated with the reproducibility of this paper.
\emph{Hence, there is a need for further investigation and initiatives encouraging AI/ML authors to follow the known reproducibility guidelines.}

\textbf{Reproducibility guidelines - beyond documentation.} The main paradigm in machine learning has shifted from classic supervised methods and hand-coded algorithms to neural models with a huge number of parameters, which are both data-intensive and computation-intensive.
\textit{The methods for communicating research results have to change accordingly, as the classic form of a scientific paper no longer describes all the necessary aspects of data, code, workflow and meta-parameters tuning in enough detail to reproduce them, both in terms of the proposed method and its experimental evaluation}.

Accordingly, the reproducibility guidelines proposed in literature related to general AI define best practices pertaining to methods, data, and experiments (see Section \ref{sec:recommendations}), but often focus on documentation of these aspects, overlooking technical issues that may render machine learning works irreproducible, particularly if deep learning models are applied. The dependencies between reproducibility and research artifacts considered by Tatman {\em et al.} \cite{Tatman2018}, Raff \cite{raff2019step}, and explained in more detail in \cite{raff2021survival} show that mathematical explanation, high-level pseudo-code and toy examples are insufficient to achieve reproducibility in machine learning.
This view of \cite{raff2019step,raff2021survival,lopresti2021} is corroborated by \cite{vincent2019,chen2022towards,pham2020variance}, that demonstrate \textit{\textbf{randomness}} and \textit{\textbf{variance}} in  experiments with deep neural networks learning, which strongly depends on the \textit{\textbf{algorithmic choices}} (e.g. neural network architecture), but also on the toolchain being applied, and the used hardware.  

\textit{\textbf{Indeterminism}} 
is an important problem in machine learning that can appear even when the original code and data are available.
It can be caused by software, e.g. third-party libraries, which might use (directly or by invoking lower-level libraries) stochastic processes \cite{Lynnerup_et_al_2020}, but also by modern hardware, as the Graphical Processing Unit (GPU) do not guarantee
to yield exactly the same results on different GPU architectures and their variants \cite{mlsys2022}.
What important, indeterminism might occur not only because of using different software or hardware components, but even due to different configuration of the same software, which can be observed for the popular TensorFlow and Keras libraries for deep learning \cite{chen2022towards}.
Unfortunately, the existing works on reproducibility address these problems only partially. 
\textit{Researchers in software engineering offer technical solutions to circumvent indeterminism \cite{navarro2020}, but usually do not investigate the improvements in machine learning reproducibility due to applying such techniques}.
Moreover, elements of randomness may appear in some of the methods themselves, especially when searching for solutions or optimizing parameters.

In the next subsection we analyse the reproducibility issues and solutions specific to deep learning, whereas the next two subsections review how reproducibility is considered in biomedical and technical applications of machine learning.
Upon these analysis we attempt to extend some of the guidelines listed in Table \ref{tab:ExperimentRequirements}
and Table \ref{tab:MethodRequirements} providing aspects specific to either DL methods or some application areas.
We relate these extended guidelines to the tools and software platforms that might be of interest to DL researchers pursuing better reproducibility in their projects.

\subsection{Reproducible deep learning}
\label{sec:deeplearning}


This subsection develops further the discussion of reproducibility issues in DL that we have identified in the literature.
It summarises recent results concerning tracking of DL experiments, dealing with details of the technical stack, and managing the assets.
The main aim of this part of our survey is to add understanding to the works on reproducibility problems in machine learning, providing a more detailed analysis of the methods for reproducible training of DL models and the techniques for dealing with randomness and indetermism, as we identified these issues as the major obstacles to wider adoption of reproducible deep learning.

\noindent\textbf{Tracking deep learning experiments.} 
A comprehensive taxonomy of tools for tracking ML experiments is described by Quaranta {\em et al.} \cite{quaranta2021}.
They address their survey to data scientists, considering that achieving reproducibility can be even more challenging in heterogeneous data science teams, when some of the members may lack software engineering expertise.
Thus, authors reviewed a large and varied set of tools for reproducible ML experiments\footnote{The full comparison is available at \url{https://github.com/collab-uniba/Software-Solutions-for-Reproducible-ML-Experiments}},
focusing on tools for data and code maintenance, but rather neglecting randomness and indeterminism.
Command-line tools, e.g. DVC \cite{DVC}, are recommended in \cite{quaranta2021} for researchers with good software engineering experience, as they are flexible and easy to integrate with Git-type version control.
If the crucial feature is the graphical comparison of experiment runs, then API-based software, e.g. MLflow \cite{MLflow} should be considered.
\emph{The analysis of this work shows, that the issues of documenting ML, and in particular DL experiments should be discussed in a broader context, considering the technical stack details, because if such details are irreproducible, they easily render the whole research work irreproducible.}


\noindent\textbf{How to deal with details of the technical stack.} 
A detailed guidance for reproducing existing deep learning models/algorithms by re-implementing the results of a published paper is given by \citet{banna2021}.
The purpose of this report is to define a process for reproducing a state-of-the-art machine learning model at a level of quality suitable for inclusion in the TensorFlow Model Garden\footnote{\url{https://github.com/tensorflow/models}}.
This work leverages the notion of an \textit{\textbf{exemplary implementation}}, 
and proposes several checklists, that serve at particular stages of engineering such an implementation:
\begin{itemize}
 \item \textit{\textbf{General checks}}: model purpose, code availability, language/framework/libraries, networks referenced.
 \item \textit{\textbf{Model and design checks}}: model architecture, model sub-networks, model building blocks, custom layers, loss functions, output structure.
 \item \textit{\textbf{Training and evaluation checks}}: dataset used, pre-processing functions, output processing functions, testing and target metrics, training steps.
\end{itemize}
A step-by-step explanation of the creation of a DL model compatible with the TensorFlow Model Garden is described on an example of a YOLO family network.
This work leverages DevOps/MLOps practices to foster reproducibility, but again treats the technical stack issues (software frameworks and libraries, specific hardware, etc.) in deep learning as secondary.

\noindent\textbf{How to train to make model reproducible.}
A very different perspective is taken by the work of Chen {\em et al.} \cite{chen2022towards},
which proposes a systematic approach to training reproducible DL models.
This approach consists of three parts:
\begin{itemize}
 \item a set of general criteria to thoroughly evaluate the reproducibility of DL models for two different domains;
 \item a unified framework which leverages a \textit{\textbf{record-and-replay}} technique to mitigate software-related randomness and a  \textit{\textbf{profile-and-patch}} technique to control hardware-related non-determinism;
 \item reproducibility guidelines which explain the rationales and the mitigation
       strategies on conducting a reproducible training process for DL models.
\end{itemize}
Chen {\em et al.} focus on the reproducibility of DL models during the training process. 
They follow the definitions used in \cite{pineau_et_al_2020},  where a particular piece of work is considered as reproducible, if the same data, same code, and same analysis lead to the same results or conclusions.
\textit{The same training process requires the same setup, which includes the same source code (including training scripts and configurations), the same training and testing data, and the same environment}.
Authors of \cite{chen2022towards} point to the lack of systematic guidelines in the literature, that take these issues into account.
They define two factors that make DL (often) irreproducible: \textit{\textbf{randomness}} in the software and \textit{\textbf{non-determinism}} in the hardware, whereas there are four types of assets to manage in machine learning in order to achieve model reproducibility: resources (e.g., dataset and environment), software (e.g., source code), metadata (e.g., dependencies), and execution data (e.g., class labels).
\textit{\textbf{Patching}}\footnote{\url{https://en.wikipedia.org/wiki/Patch_(computing)}} methods that apply a set of changes to a program or its supporting data could be necessary to fix randomness issues in the software. 
Prior work shows that other assets should not be managed with the same tools (e.g., Git) as used to manage source code.
Hence, version management tools (e.g., MLflow \cite{MLflow}, DVC \cite{DVC}) are recommended for managing DL assets.
Interestingly, Chen {\em et al.} show that the existing general guidelines \cite{mitchell_modelcards_2019} are insufficient in some scenarios of deep learning, due to the undocumented  software dependencies.
The paper provides general guidance, that should be followed in order to achieve reproducibility in deep learning, but defines also a more rigorous procedure, that consists of a number of steps and includes special tools (record-and-replay), which makes it possible to achieve reproducibility in DL tasks:
\begin{enumerate}
 \item \textit{\textbf{Documentation frameworks}} such as Model Cards \cite{mitchell_modelcards_2019} should be used to document training,
       documenting also software dependencies.
 \item \textit{\textbf{Tools}} such as DVC\footnote{\url{https://dvc.org/}} or MLflow\footnote{\url{https://mlflow.org/}} should be applied to manage the experimental assets,
       using virtualisation techniques to provide a complete runtime environment,
       which mitigates the risks of non-determinism.
 \item \textit{\textbf{Appropriate metrics}}, general or domain specific should be used to document the model evaluation criteria.
 \item \textit{\textbf{Pre-set random seeds}} or a record-and-replay technique has to be applied in order to mitigate sources of randomness in the software. 
 \item \textit{\textbf{Patching methods}} should be considered to mitigate non-determinism from hardware.
 \item \textit{\textbf{Unsupported non-deterministic operations}} should be documented, and replaced by their deterministic alternatives. 
\end{enumerate}

\noindent\textbf{Managing the assets in deep learning.}
The management of assets in DL experiments can be also accomplished using dedicated software, such as dtoolAI \cite{hartley2020},
which allows to automatically capture data inputs and model hyper-parameters while training a DL model, and to distribute those
metadata with the model.
What is more, dtoolAI\footnote{\url{https://github.com/JIC-CSB/dtoolai}} allows attaching unique URIs (Universal Resource Identifiers) to datasets hosted by cloud services,
allowing datasets to be uniquely identifiable.
This is a Python software package integrating with the Keras library and based on dtool \cite{olsson2019}, an application programming interface (API) and set of tools for management of heterogeneous data. 

A different approach to managing DL assets is demonstrated by DeepForge \cite{Broll2017DeepForge}, a platform for deep learning that enables rapid development of DL models in a cloud-based infrastructure. 
It leverages the concept of model integrated computing \cite{modelintegrated1997} and provides a hybrid text-visual programming platform. 
DeepForge\footnote{\url{https://deepforge.org/}} facilitates reproducibility of DL experiments using automatic versioning during development, and enabling versioning both the code and the binary artifacts (e.g. data and trained models) for the given experiment. 

\noindent\textbf{Dealing with randomness and indeterminism.}
Whereas managing the provenance of a computational objects is a cornerstone of reproducibility, the issues of randomness and indeterminism from both software and hardware also need to be considered if we want to obtain a fully reproducible DL experiment, which yields quantitatively adequate results when repeated by others \cite{chen2022towards}.
Thus, it is necessary to investigate closely the origins of the problem \cite{pham2020variance,mlsys2022}.
\textit{As shown by the experimental results in \cite{pham2020variance}, identical training runs of a deep learning method, with same algorithm, network architecture, and training data still can produce different models characterised by different accuracy}. This is caused by a number of issues, that may be divided into two categories, named in \cite{pham2020variance} algorithmic factors and implementation factors. The same taxonomy is used in \cite{mlsys2022}.

The \textit{\textbf{algorithmic factors}} are related to model design choices, and they appear when stochastic components are utilised to improve model accuracy and training efficiency.
Examples are: \textit{\textbf{random initialisation of weights}}, \textit{\textbf{stochastic layers}} (e.g. dropout) in the network, \textit{\textbf{ordering of data shuffling}}, and \textit{\textbf{stochastic data augmentation}}.

The \textit{\textbf{implementation factors}} are related to indeterminism introduced by either software or hardware components of the technical stack.
Commonly used deep learning libraries, such as TensorFlow and PyTorch, perform data pre-processing in parallel to improve speed, but this approach can change the order of training data.
Differences in training data ordering result in different variance caused by floating point operations.
Moreover, different runs of a DL method using such libraries sometimes use different elementary operations on the GPU (from cuDNN lower-level library), which introduces differences between the resulting models.
The GPU itself can introduce implementation noise due to different orders in the sequence of floating-point operations that result in differences in rounding error propagation \cite{jezequel2015}.
This phenomenon is exemplified by Morin and Willets \cite{morin2020gpu}, who found that randomness in the results of image classification by ResNet models
is dominated by non-determinism from GPU rather than by algorithmic factors.
According to \cite{mlsys2022} these problems are observable across a wide range of parallel processing units, including TPUs.
\textit{Moreover, reproducibility could be compromised if the model training is not convergent, or if performance depends on the size of testing data}.
Liu {\em et al.} \cite{grundy2022} found that low convergence of the training process or testing on data of considerably different scale than those used while training can lead to DL model performance that is difficult to reproduce.

Because most of the algorithmic factors of irreproducibility are controlled by generators of pseudo-random numbers, one can avoid these factors in a reproducible deep learning experiment just setting the random seeds at the beginning of each run \cite{madhyastha2019}.
Note that the pseudo-random behaviour often needed for training efficiency will be present within a single run \cite{pham2020variance}.
The non-stochastic aspects of a deep learning method that can compromise reproducibility, and thus have to be chosen carefully are hyper-parameters \cite{lucic2018gan}, and the choice of the activation functions \cite{shamir2020distil}.

Elimination of the implementation factors that compromise reproducibility requires to take a number of steps, which depend on the DL task at hand and the software/hardware setup being used.
Pham {\em et al.} list some general recommendations for these steps:
\begin{itemize}
 \item The use of \textit{\textbf{multiple processes}} that cannot guarantee data order should be avoided.
 \item If the used libraries apply \textit{\textbf{autotune mechanisms}}, then deterministic implementations of primitive operations
       should be chosen setting appropriate configuration variables.
 \item If \textit{\textbf{GPU computations}} are used, then techniques that ensure deterministic execution of operations should be considered.
\end{itemize}
Zhuang {\em et al.} explain the impact of particular choices in the DL toolchain on reproducibility in more detail, considering also the cost of controlling the implementation factors in deep learning.
Techniques facilitating reproducibility of GPU workloads, that may be useful for DL tasks, have been proposed by computer architecture researchers \cite{jooybar2013,chou2020atomic}.

A major role among the software engineering solutions that help to overcome the technical causes of irreproducibility is played by the \textit{\textbf{containerisation techniques}}, such as Docker \cite{merkel2014}. 
These techniques make it possible to more easily create reproducible software environments, for example by creating a Docker image based on a repository path or URL \cite{forde2018repo}. 
Although standard dockerfiles are helpful, they do not solve a number of problems related to the implementation factors of reproducibility, including problems specific to the operating system and hardware platform.
An example of software technology that can solve these problems is described by Navarro {\em et al.} \cite{navarro2020}.
Their paper gives an extensive discussion of the sources of irreproducibility in the Linux API
calls and the x86-64 ISA instructions, and describes DetTrace, a reproducible container abstraction for Linux implemented in user space.
All computation that occurs inside a DetTrace container is a pure function of the initial filesystem state of the container.
It is shown that DeTrace can be used to render deep learning tasks reproducible: to check the reproducibility of TensorFlow workloads, authors recorded the value of the loss function at each step during training.
These values are irreproducible when running natively, even with serialized TensorFlow, due to, e.g., randomization of the training set.
DetTrace \cite{navarro2020} renders these workloads reproducible without any code changes.
\textit{Reproducible containers can be used for a variety of purposes, but apparently, it may be of interest also for those, who want to conduct reproducible deep learning research}.
DetTrace seems to solve the technical side of DL reproducibility (implementation factors) without requiring any hardware, operating system or application changes, but with some time overhead that depends on the DL task itself.

\subsection{Reproducibility in artificial intelligence for biomedical applications} 
\label{sec:medical}
The recent breakthroughs in AI and ML have demonstrate a great potential in supporting many biomedical decision problems, such as disease risk prediction \cite{GIORGI2022} 
or diagnostic image analysis \cite{Bertelli2021}. Not by chance, a new paradigm termed \emph{“High-performance medicine”} has been anticipated by Eric Topol, who envisaged “the convergence of human and artificial intelligence” as a promising perspective \cite{Topol2019}.  
However, despite the great expectations and promises of AI in the medical field, the actual impact of AI-powered tools on clinical routines appears anticipatory, as many issues still prevent their full uptake in real-world practice. Among these issues, reproducibility is among the most critical ones, coupled with reliability, accountability, liability, and overall trust \cite{future-ai}. The publication-based dispute about the reproducibility and transparency of an AI biomedical solution for the screening of breast cancer is paradigmatic of the cogency of the topic \cite{McKinney2020,Trasp2020}. 

An evidence of this raised interest 
is the noteworthy share of papers retrieved by our queries that focused on the biomedical field (i.e., 18 out of 81, accounting for the 22\% of the papers, though not all relevant for our purposes). These papers 
addressed 
the recently surging demand for transparency and trust, though it is worth noting that reproducibility is an old quest for the results coming from the statistical and ML analyses of medical data \cite{Carlson2012,Hegselmann2018ref17raw57,Johnson2017ref24raw57}.

As it happens in the broader scientific community, the terminology behind reproducibility varies a lot within  the biomedical application domain, so demonstrated the papers that we retrieved as described in more details in the supplementary material in Section \ref{sec:term_biomedic}. Among them, many work referred to a general concept of reproducibility, with the goal to raise awareness on its importance \cite{Trasp2020,raw48}. Other works made an explicit reference to one of the definitions of reproducibility we have overviewed in the previous Section 3 \cite{raw81,raw65}; others introduced additional notions so to take into account the peculiarities of the biomedical domain \cite{raw57}, and some others considered the whole spectrum of reproducibility dimensions (i.e., the four quadrants) as introduced in Section 3 \cite{raw59}. The most relevant works debated the specificity of the biomedical domain and provided recommendations on how to cope with it \cite{raw57,raw59,raw81}. In the following, we will first overview 
the criticalities that affect reproducibility in the biomedical domain, and then we discuss the guidelines or recommendations that are provided by the most relevant papers retrieved.
\\
\\
\textbf{Criticalities of AI reproducibility in biomedical applications}. 
AI reproducibility in the biomedical field is affected by several factors. Some of them are those common to all scientific domains and comprise the lack of a complete documentation of the development process, the training pipeline and the development choices done as well as the lack of a suitable management of the randomness and indeterminism characterizing the solutions based on DL \cite{raw81, raw18}. The lack of code sharing is even more critical, as it is a less common practice in the biomedical community, where the use of open-source repositories is traditionally not common \cite{raw18, raw57}. Although awareness on this issue is increasing, still a low share of the published papers releases the code or any supplementary materials \cite{raw18}. Further to this, other challenges appear to be peculiar of the biomedical domain as it has been highlighted by many of the retrieved papers \cite{raw81,raw57, raw18,raw29}. We summarize these challenges as follows. 

\begin{itemize}
\item \textbf{\emph{Difficulties in sharing data}}: biomedical data are privacy sensitive and their sharing policies are subject to specific authorization by individual owners and clinical institutions. Privacy preservation and suitable anonymization techniques should be applied to avoid any leakage of sensitive information. Nonetheless, these are still partially used and sometimes they may hinder the utility of the shared data \cite{raw57,Dwork_Ullman_2018}. Consequently, very few publications share the datasets used for training and testing their AI models. This highly affect the repeatability and reproducibility dimensions \cite{raw81,raw18,raw57}

\item \textbf{\emph{Lack of heterogeneous and multi-institutional data}}: most of the works on AI in biomedical applications make use of ad hoc datasets, often collected 
by a single clinical institution as retrospective data. These datasets unfortunately exhibit a low capacity to represent the large variability of real-world data. 
As explained for instance for the medical imaging data \cite{Castro2020}, variability may be caused by the heterogeneity and differences among clinical institutions, in terms of different patient population, clinical protocols implementation, operators’ practices as well as the data acquisition procedures and equipment. 
Additional 
variability comes from the heterogeneity of individual patient’s characteristics, in terms of anatomical, functional and response properties. 
Moreover, as the clinical knowledge evolves, also the data acquisition protocols, the clinical guidelines and standards of care, the reference ground truth, and the acquisition devices change over time, thus causing the data and the biomedical decision questions to experience variability and drift over time \cite{Sammut2010}. This variability strongly affects 
replicability and generalizability dimensions of reproducibility. 
Specific measures to cope with this condition should be adopted along the entire life-cycle of AI development \cite{future-ai}, and particularly at the moment of data selection, by gathering heterogeneous datasets, acquired from diverse institutions and representing a large population share \cite{raw57}. 

\item \textbf{\emph{Lack of standardized evaluation policies}}: the variability introduced in the previous point should be managed also from a statistical standpoint when presenting the outcomes of an AI model. The statistical evaluation based for instance on suitable stratified cross-validation techniques that avoid any forms of data leakage between training, test and validation (e.g., in medical imaging analyses by separating data at patient level and not at slice-level) are not often adopted. This issue mainly affects the development process, but, when coupled with the lack of documentation, it strongly affects the possibility to reproduce the claimed results. As recommended by Wojtusiak \cite{raw57}, statistical model-quality measures, some characteristics of the learning algorithms, including learning curves, attribute selection curves, and hyper-parameter tuning curves are not often reported in the published papers, while they should be included accompanied by relevant statistical measures.

\item \textbf{\emph{Lack of reporting standards}}: 
an extensive and transparent reporting of a study, experiment or trial have been always considered as a key issue to adequately assess any risk of bias as well as the safety, efficacy and potential uptake in clinical practice of any prediction models or clinical interventions \cite{McErlean2022}.  Standardized reporting strategies have been devised for this purpose, such as the 
TRIPOD  \cite{tripod}, the 
CONSORT \cite{consort} and the 
SPIRIT \cite{spirit} checklists. Nevertheless, the current version of these checklists does not include items specific to AI and ML methods. Most of them are currently working to fill this gap with dedicated releases, such as the TRIPOD-AI \cite{Collinse048008}, the 
CLAIM checklist \cite{claim}, and its minimum information MI-CLAIM version \cite{mi_claim}, as well the CONSORT-AI and the SPIRIT-AI \cite{spirit_AI_consort_AI}. Currently there are no broadly accepted guidelines and no standardized reporting used in practice \cite{additional_Collin2022}. 
\end{itemize}

\textbf{Recommendations and guidelines}. In the last few years, as testified by the results of our queries, some authors have published recommendations and guidelines to promote the reproducibility of AI solutions in the biomedical domain. Also scientific journals have recommended authors to fill ad-hoc checklists to  verify the reproducibility of their work \cite{IJMEDI_checklist}. In many cases, the scope is broad and focuses on the overall transparency, scientific excellence and trust of AI development processes \cite{Luo2016,future-ai,claim}. Indeed, for some issues, such as traceability, soundness of the approach or precise documentation, it is difficult to set a border between what pertains only reproducibility and what goes beyond. For instance, the explicit identification of the authors of the work (or part of it) is not necessarily mandatory for reproducibility, as far as the technical documentation is sound and complete to enable experiment reproduction. Instead, it may be mandatory and needed for traceability and accountability purposes.  The more structured guidelines consist in either a checklist of items to be verified \cite{claim,future-ai}, or in a list of criteria to be met or questions to be answered \cite{raw59, Stevens2020,Vollmer2020raw40}. In other cases, general recommendations are provided \cite{raw57}. 

Overall, the most common recommendations touch upon the following issues:
\begin{itemize}
\item \textbf{\emph{Process documentation}}: focusing on the need to annotate and document any data, processing steps and choices done during development as well as the hardware and software
\item \textbf{\emph{Standardized reporting}}: focusing on how to standardize a detailed report of methods and results. This moves the previous issue further, by proposing structured checklists and reporting items
\item \textbf{\emph{Resource sharing}}: promoting the appropriate sharing of data, code and/or AI models details (such as weights, input and output sets, training hyper-parameters)
\item \textbf{\emph{Development strategies}}: recommending scientific excellence based on statistical soundness and strategies to foster replicability and generalizability of AI models (e.g., via data representativeness).
\end{itemize}

Some of these recommendation categories are cross-sectional to the criticalities described in the previous section (i.e., process documentation and resource sharing) and are common to almost all the work on reproducibility. Others address specific criticalities (i.e., standardized reporting) or regard the technical measures to be adopted to cope with all the mentioned criticalities.

Most of the recommendations provided strongly overlap with the general recommendations we have overviewed in the previous sections, namely Section \ref{sec:guidelines} and Section \ref{sec:reasoned_guidelines}. Some more specific recommendations mainly focus on the types of information to be specified when documenting the dataset (e.g., inclusion \& exclusion criteria, population statistics, selection criteria), and the development strategies towards statistical soundness (e.g., collection of multi-institutional datasets or model calibration). For the sake of completeness, the most relevant recommendations are overviewed in the following, grouped per issue they address. They are also further summarised in Section 6 within Tables \ref{tab:DNNRequirements2} and \ref{tab:DNNRequirements3}. 

As for \textbf{process documentation}, almost all the papers retrieved strongly pushed authors to accurately document the detailes of their work, as documentation is a key enabler of reproducibility \cite{raw48, Trasp2020, raw57, raw59, raw81, raw18, raw39, Vollmer2020raw40}. More specifically, Renard et al. proposed to characterize and document any forms of uncertainty and variability in the AI development, especially in the case of DL \cite{raw81}. Namely, they proposed to document the dataset used, the optimization and process, the hyper-parameter choices, the DL architecture, the middleware and the infrastructure used. 

As far as \textbf{standardized reporting} is concerned, Stevens et al. \cite{Stevens2020} presented reporting criteria split into four main categories: study design, data sources and processing, and model development and validation. Vollmer et al. \cite{Vollmer2020raw40}, discussed a framework consisting of 20 criteria (questions) intended to guide ML and statistical research, split into six categories: inception, study, statistical methods, reproducibility, impact evaluation and implementation. Reproducibility and quality of work was also addressed in the context of clinical trials \cite{Wicks2020}. The work by McDermott et al. \cite{raw59} listed ten machine learning reporting items covering the following phases: experimental design, statistical model evaluation, model calibration, top predictors, sensitivity analysis, decision analysis, global model explanation, local prediction explanation, programming interface, and source code.

When considering \textbf{resource sharing}, Beam et al. proposed to share data in dedicated biobanks in order to guarantee suitable data governance \cite{raw18}. In case data sharing would raise privacy concerns, they suggested to sharing data according to a “walled-garden” approach whereby reviewers might be provided access to a private area and use the data for reproducibility analyses during the review period. Code release on any of the most common repositories (e.g., Github, Bitbucket, or GitLab) is the most common recommendations \cite{Trasp2020, raw57, raw59}, also mentioning containerization to avoid dependences to software environment. 

As far as \textbf{development strategies} are concerned, the most complete recommendations came from Wojtusiak \cite{raw59} and McDermott et al. \cite{raw57}. They proposed to:
\begin{itemize}
\item integrate multi-institution datasets, in order to ensure the heterogeneity of data and representativeness of the dataset used to developed the AI solutions
\item prospectively collect data, in order to avoid developing AI models that might likely not generalize on real-world prospective data
\item develop new privacy-preserving analysis techniques, to ensure data and can be shared without disclosing any sensitive information while retaining the relevant information 
\item adopt rigorous statistical approaches for the development and evaluation of AI models, based on statistical best practices, in particular when considering model comparisons, model calibration, identification of top predictors and sensitivity analyses.
\end{itemize} 

Besides the aforementioned types of recommendations, some authors addressed their suggestions also to regulatory bodies, encouraging the enactment of specific policies for AI study in the medical field. For instance, McDermott and co-authors also suggested to adopt a \emph{pre-registration} policy for AI-based studies, meaning that the experimenters or authors should report to regulatory bodies their goal and planned preprocessing/modelling scheme before they run any experiments \cite{raw57}. This would intend to avoid intentional or unintentional statistical frauds.

\subsection{Reproducibility of ML methods for physical artificial intelligence} 
\label{sec:physical}

\textit{\textbf{Physical AI}} refers to using AI techniques to solve problems that involve direct interaction with the physical world, for example by observing the world through sensors or by modifying the world through actuators. 
The data is generated mostly from physical sensors, but might be combined with other sources, such as databases, the Internet or direct user input. Actuation may range from support to human decisions to managing automated devices (e.g., traffic lights, gates) and actively directing robots, autonomous cars, drones, etc. \cite{ai4eu2020}.

Whereas physical AI largely uses the same methods and algorithms that are already proven in other application areas, including the growing share of deep learning based solutions, the specific sources of data and the forms of output make this area particularly challenging with respect to reproducibility.
Science and engineering in general are dealing with physical experiments routinely, and there are established guidelines and protocols that allow researchers to keep that random factors under scrutiny.
The results are experimental outcomes that are at least statistically reproducible and comparable.

\textit{Therefore, we can think about reproducibility in physical AI in terms of casting the recommendations for reproducibility we devised specifically for AI and machine learning onto the well-established rules of conducting reproducible experiments in general science and engineering}.
However, there are problems specific to physical AI, particularly concerning robotics, that need to be tackled in order to ensure a common understanding of reproducibility and its standards among researchers.
The number of publications we were able to survey that are concerned with reproducibility in physical AI seems to be much smaller than the number of those concerning other application areas, e.g. medical, while different aspects of physical AI are represented to a much different extent in this literature.
The following paragraphs of this subsection deal with reproducibility issues in Reinforcement Learning and Deep Reinforcement Learning, which are the most promising approaches to learning in physical AI. 
We tackle also the problem of comparing new results to state-of-the-art baselines, as this is particularly problematic if physical systems are involved, while the lack of reproducibility makes it often impossible to asses the importance of newly published results. 
Problems in achieving reproducibility when ML/DL is applied to two specific categories of physical systems,
namely, if we deal with physical robots, or we try to obtain reproducible simulations of physical systems, 
are considered as being beyond the main scope of the this survey.
However, as these issues might be interesting to readers applying modern AI solutions in robotics, they are covered in section C of the supplementary material available with the electronic version of this paper.

\noindent\textbf{Reproducibility of reinforcement learning experiments.}
A well represented aspect is \textit{\textbf{reinforcement learning}} (RL) \cite{henderson2018renforcement}, which recently achieved successes in solving complex tasks of physical AI, particularly in robotics \cite{roboticsrl}. 
These achievements are also attributed to enhancing the RL paradigm with deep learning, that gave rise to the branch of \textit{\textbf{deep reinforcement learning}}.
Unfortunately, reproducibility in RL is even harder to obtain than in other areas of ML/DL, because a RL learning agent needs to 
interact on-line with an environment by taking actions and receiving feedback of their consequences (a reward). 
\textit{This manner of gathering the learning data makes RL experiments prone to additional non-determinism due to stochastic environment \cite{islamreproducibility2017} and physical issues, such as delayed sensor readings \cite{Lynnerup_et_al_2020}}.
Khetarpal {\em et al.} \cite{Khetarpal2018} were one of the first who studied the importance of reproducibility in reinforcement learning. 
\textit{They pointed out the distinction between reproducibility of the algorithm and reproducibility of the evaluation procedure, focusing mainly on the latter issue}.
An evaluation pipeline for (deep) RL that could be standardised was proposed in \cite{Khetarpal2018}, which was the basis for the procedures used in \cite{Lynnerup_et_al_2020}.
Whereas RL is often used in simulated environments or simplified worlds, such as arcade games, Lynnerup {\em et al.} \cite{Lynnerup_et_al_2020} tackled the problem of reproducing deep RL algorithms on real robots.
They used the SenseAct\footnote{\url{https://github.com/kindredresearch/SenseAct/}} framework with an Universal Robots' manipulator arm to solve a benchmark task in which the robot has to reach target points in a 2D plane with its end-effector.
This paper contributed a demonstration of a rigorous method for documentation of parameters with experiment configuration files, and a statistics-based evaluation of common RL baseline algorithms on real-world robots. 
Moreover, \cite{Lynnerup_et_al_2020} comments extensively the sources of non-determinism in machine learning, deep learning, and those specific to deep RL: 
\begin{itemize}
 \item \textit{\textbf{environment}} -- when dealing with physical systems, sensor delays, network/interface delays and other uncontrolled issues might be the prevailing source of indeterminism; 
 \item \textit{\textbf{initialisation}} of the neural networks’ weights in deep RL must be controlled to ensure reproducibility; 
 \item \textit{\textbf{sampling randomly}} from the training data and from replay buffers (minibatch sampling).
\end{itemize}
In the light of these findings and the experimental results \cite{Lynnerup_et_al_2020} proposes a list of recommendations for reproducible deep RL experiments, which are basically in-line with those devised by others for general-purpose DL. 
\textit{Lynnerup {\em et al.} suggest also frequent code reviews to ensure the integrity of code prior to conducting experiments, and acknowledge that reproducing RL results on a real robot introduces many technical issues which are difficult to debug, while they may contribute to uncontrolled randomness in the experiment}.

The impact of non-determinism on reproducibility in deep Q-learning \cite{mnihnature} was investigated in \cite{Nagarajan2018}, with a conclusion that a non-deterministic implementation of reinforcement learning may not reproduce results of similar quality to published results, solely due to the large variance between runs.
Investigating the sources of non-determinism, 
Nagarajan {\em et al.} \cite{Nagarajan2018} point out to issues related to \textit{\textbf{initialisation and the use of GPU}}, which is consistent with the results reported for deep learning in general, but also emphasize that in RL the agents typically employ a stochastic policy during learning, i.e. the agent’s action is drawn from a non-degenerate distribution over the available actions. 

\noindent\textbf{Reproducibility for comparing to baselines and competitors.}
Achieving reproducibility is important when a researcher wants to compare her/his results to the existing baselines.
When physical AI is considered, this factor becomes crucial, because small implementation decisions make a big difference, as demonstrated in \cite{Julian2020} for the case of multi-task RL (MTRL).
This paper shows that small implementation details can create absolute performance differences of up to 35\% on meta- and multi-task RL benchmarks for a given algorithm.
\textit{This results in a statistically-significant variance in the performance of the investigated algorithm that can exceed the reported performance differences between this algorithm and the baseline, thus confounding performance improvement claims of the researchers}.
Julian {\em et al.} \cite{Julian2020} emphasize the role of widely-disseminated \textit{\textbf{reference implementations}} and consistent \textit{\textbf{evaluation protocols}}, providing multi-seed benchmark results on the available benchmarks for several most popular meta-RL and MTRL algorithms.
The problems related to reproducibility of robotics research obviously mount up with the increasing complexity of the robotic system.
In this context \cite{swarmrob2018} proposed the SwarmRob toolkit\footnote{\url{https://iot-lab-minden.github.io/SwarmRob/}},
which deals with the problem of sharing experimental artifacts in multi-robot systems. 
SwarmRob leverages virtualisation of robotics applications isolating the artifacts of experiments in containers that can be easily shared with others, facilitating reproducibility.


\subsection{Recommendations for improving reproducibility in machine learning and its applications}
\label{sec:MLrecommend}

The extended discussion of the reproducibility issues in machine learning based on our literature studies revealed that the techniques and tools of modern machine learning, particularly deep learning and reinforcement learning, introduce additional factors that  compromise reproducibility.
Whereas the general recommendations for reproducibility we summarised in section \ref{sec:recommendations} are still valid for all areas of machine learning, the discussion shows that DL, RL and their emerging applications in medicine and physical AI require to take particular care about some specific factors. 
Therefore, in Table \ref{tab:DNNRequirements2} we re-visit some of the general guidelines for reproducibility in the light of these factors, providing for each of the issues a very brief recommendation devised on the basis of our analysis, the area of ML where the issue is most pronounced, and references to example papers that offer a more detailed treatment of the given problem. 

\begin{table}[th]
\scriptsize
\begin{tabular}{lp{0.4\linewidth}p{0.1\linewidth}l}
\textbf{Feature} &
  \textbf{Recommendation} &
  \textbf{Application context} &
  \textbf{Source guideline} \\
  \hline  
Hyper-parameters & Use tools that automatically capture data inputs and model hyper-parameters while training a DL model. Store the training parameters and the input data together with the AI model (e.g. dtoolAI) & All DL areas & \cite{hartley2020}\\
Measure & Document the measurement metrics, which is essential for reinforcement learning. & Deep RL & \cite{Lynnerup_et_al_2020} \\
Workflow &
  Use documentation frameworks to document training, documenting also software dependencies.
  Avoid algorithmic factors of irreproducibility setting the random seeds at the beginning of each run
  & All ML areas & \cite{chen2022towards}, \cite{madhyastha2019}  \\
Workflow execution & Employ automatic versioning tools enabling versioning both the code and the binary artifacts (e.g. Deep Forge).  & All DL areas & \cite{Broll2017DeepForge} \\
Data collection, annotation and quality & Document carefully the data acquisition process for physical AI systems. Annotate the training data with metadata and give them a persistent identifiers (e.g., URIs). & Biomedical applications, Embodied agents, Deep RL &  \\
Data collection, annotation and quality & Report the inclusion and exclusion criteria for the considered cases. Provide descriptive statistics of data. Describe the acquisition protocols and parameters & Biomedical applications &  \cite{raw59}\\
Preprocessing cleaning and labelling & Provide a detailed description of the calibration procedures that were applied to the physical system that produced te data & Embodied agents, Deep RL &  open issue \\ 
Preprocessing cleaning and labelling & Describe the expertise of data annotators and the process to reach consensus among them. Describe and document the tools and the parameters set for annotating the data. Describe any pre-processing step implemented for enhancing the quality of the data or harmonize different datasets & Biomedical applications & open issue \\
Raw data & If this is feasible, make the raw data available along with the pre-processed/labelled/cleaned data & Embodied agents, Biomedical applications & open issue \\
Preprocessing software & If possible use commodity, open-source software for data manipulation, documenting the process.
If proprietary software is used, make it available and document at the same (high) level of details as the entire project & All DL areas & open issue \\
\end{tabular}
\caption{General recommendations from Table \ref{tab:ExperimentRequirements} or Table \ref{tab:DataRequirements} that need to be extended and/or clarified in the context of deep learning, DL/RL in embodied agents or biomedical applications. Note that items in the {\bf Feature} column strictly match their counterparts in the first columns of Table \ref{tab:ExperimentRequirements} or Table \ref{tab:DataRequirements}}
\label{tab:DNNRequirements2}
\vspace{-0.6cm}
\end{table}

Our survey reveals also that there is a group of reproducibility issues specific to deep learning, reinforcement learning, biomedical applications or physical AI systems that are largely overlooked in the general recommendations.
These issues are related to the specific algorithmic or implementation factors in DL/RL or to the way data collected from physical systems (either off-line or on-line) is treated for training of ML systems. 
Identifying these problems in the surveyed literature (often very recent conference or workshop papers) and confronting them to the list of general guidelines we identify gaps in the understanding and awareness of the AI community with respect to reproducibility.
Table \ref{tab:DNNRequirements1} summarises the recommendations that are specific to DL/RL methods and their application areas. Analogously as in the previous table, we provide for each issue a brief recommendation to improve the reproducibility level, define the area where these recommendations are most useful or important, and list the papers that consider the given problem and propose the recommendations. 

\begin{table}[th]
\scriptsize
\begin{tabular}{p{0.25\linewidth}p{0.4\linewidth}p{0.1\linewidth}l}
\textbf{Issue} &
  \textbf{Recommendations} &
  \textbf{Application context} &
  \textbf{Source guideline} \\
  \hline
Randomness & Pre-set random seeds or use a record-and-replay technique to eliminate sources of randomness & All DL areas & \cite{chen2022towards} \\
Algorithmic indeterminism & Avoid the use of stochastic layers in neural networks, random initialisation of weights,
random ordering of data shuffling, and stochastic data augmentation & All DL areas & \cite{mlsys2022} \\
Implementation indeterminism (software) & Chose deterministic implementations of primitive operations, avoid autotune mechanisms in libraries, avoid the use of multiple processes that cannot guarantee data order, use dockerization techniques & All DL areas & \cite{chen2022towards} \\
Implementation indeterminism (hardware) & Eliminate unsupported non-deterministic operations, use techniques facilitating reproducibility of GPU workloads, use advanced dockerization, such as DetTrace  or apply a profile-and-patch technique & All DL areas & \cite{navarro2020} \\
Comparison to baselines and previous results & Use reference implementations and consistent evaluation protocols, use benchmarks that support reproducibility (e.g. ACRV, EGAD in manipulation)
& Deep RL, embodied agents, biomedical applications & \cite{Nagarajan2018} \\
Evaluation procedures in RL & The evaluation pipeline for reinforcement learning, including deep RL, should be standardised in order to obtain reprodcucible results & Deep RL & \cite{Khetarpal2018} \\
Using public datasets of sensor data & When documenting data provenance, describe the methods and implementation stack for any pre-processing/transformation steps executed on the data. Document the version of dataset being used (e.g. using DOI), particularly if anything was changed with respect to the originally published one &  Embodied agents, biomedical applications &  open issue \\
Using simulated environments & Consider trade-off between sim2real randomisation and reproducibility. Whenever possible use a reproducibility-aware simulator, such as Gym-Ignition, Isaac Gym & Embodied agents, Deep RL & \cite{ferigo2020}
\end{tabular}
\caption{Recommendations specific to deep learning. Note that items in the {\bf Issue} column are different from the features identified in Table \ref{tab:ExperimentRequirements}, Table \ref{tab:DataRequirements}, or in the extended versions in Table \ref{tab:DNNRequirements2}}
\label{tab:DNNRequirements1}
\vspace{-0.6cm}
\end{table}

\begin{table}[th]
\scriptsize
\begin{tabular}{lp{0.4\linewidth}p{0.1\linewidth}l}
\textbf{Issue} &
  \textbf{Recommendations} &
  \textbf{Application context} &
  \textbf{Source guideline} \\
  \hline
Data selection & Integrate multi-institutional datasets. Prefer when possible prospective data. Always consider to validate models on prospective data & Biomedical applications & \cite{raw57} \\
Data curation & Adopt privacy-preserving analysis techniques, in order to enable data sharing & Biomedical applications & \cite{raw57} \\
Model evaluation & Perform model calibration and report calibration curves and calibration measures. Perform global sensitivity analysis for attributes, report sensitivity plots and analyze if models are stable. Perform global sensitivity analysis for continuous and discrete attributes and report sensitivity plots and analyze if models are stable & Biomedical applications & \cite{raw59} \\
\end{tabular}
\caption{Recommendations specific to the development of AI models in biomedical applications. Note that items in the {\bf Issue} column are different from the features identified in Table \ref{tab:ExperimentRequirements}, Table \ref{tab:DataRequirements}, or in the extended versions in Table \ref{tab:DNNRequirements2}}
\label{tab:DNNRequirements3}
\vspace{-0.6cm}
\end{table}




\section{Final remarks}
\label{sec:conclusion}

Reproducibility is one of the key dimensions that concur to create trustworthy and reliable AI and ML. It becomes paramount when demonstrating scientific outcomes or methods that result from experimental processes, not necessarily supported by known theories or models, as it happens with current data-science and learning methods. 

Despite the awareness of this and the appearance of many works on reproducibility in recent years, the take-up of reproducibility-enabling practices in practical implementation is still insufficient. In particular, this applies to the most advanced ML and DL methods, which are at the core of the most recent successes of AI. As we have extensively debated in the previous sections, several conditions complicate those methods' reproducibility. They include: the complexity of models in terms of architecture, the high number of parameters to be tuned, the optimization strategies needed to make them perform as expected, the inner peculiarities of the stochastic processes, the characteristics of the data used to their training process and all the pre-processing steps to curate and prepare these data, the technical peculiarities of the underlying tool-chains, and the dependencies to hardware and software platforms. 

Nevertheless, the more technical issues that hinder reproducibility in modern ML are often overlooked in the general literature, which is focused more on process documentation and code sharing. They are treated only in more specific technical papers, as we have analyzed in more details in Section \ref{sec:learning}. 
Moreover, specific application scenarios, such as the presented biomedical and physical AI fields, show peculiarities in terms of sensitive data sharing, data acquisition indeterminism, statistical soundness of results, evaluation procedures and detailed reporting, which further complicate the reproducibility setting.
As we attempted to show in our paper, all these issues require more systematic approaches with solutions that touch upon best practices, new requirements, quality standards and configuration choices along the technical stack to better model the development process of ML methods and their reproducibility.

In our survey, we wanted to broaden the debate over reproducibility of AI and ML. We also attempted to homogenize and map the various guidelines and recommendations, making explicit the technical or procedural means to implement them, namely via metadata, platforms, or the scientific material, paper, or report.  Moreover, we strongly advocated taking into account the additional factors that compromise reproducibility in the field of ML, DL and RL and critical application scenarios. In this respect, we dedicated the last section of this paper to propose new specific recommendations.

As a result of this process, we realized there are still several open issues that need to be taken into account for comprehensively enabling reproducibility. The most relevant ones pertain to:

\begin{itemize}
\item the limited functionalities of existing development platforms with respect to many of the presented recommendations, especially when considering procedures to handle the randomness and the variability of the training process in DL; 
\item the insufficient standardization of metadata usage. Current proposals of metadata are still far from well-documenting AI experiments, as we discussed in Section \ref{sec:recommendations};
\item the limited practice in containerizing the developed applications and sharing the entire digital artefacts (in case with the development notebooks); 
\item the lack of standardized evaluation procedures and reporting, which would include the comparison with baselines, benchmarks and competitive models, especially in the field of RL, biomedical applications and physical AI;
\item the difficulties in sharing complete data in some application areas, such as the discussed biomedical field, due to suitable anonymization techniques;
\item the open challenge of interpreting the inner functioning and the generalization capacity of complex ML, DL and RL models.
\end{itemize}

Considering the key role that reproducibility plays in increasing the overall transparency and accountability of AI and ML, we believe that further initiatives are in need to encourage researchers, authors, and industrial actors to comply with the proposed guidelines. The approach should be overarching and comprehensive, not limited to documentation and resource sharing, as it should also encompass best practices at the technical and implementation level. 

\begin{acks}
This research was partially supported by the two EU H2020 Projects TAILOR (GA 952215) and ProCAncer-I (GA 952159). 
\end{acks}

\shortcites{key-list}
\setcitestyle{compress}

\bibliographystyle{ACM-Reference-Format}
\bibliography{sample-base}

\appendix

 \begin{screenonly}

\pagebreak

\section{Methodology of the review}
\label{sec:method}

This article is written following a \textbf{\textit{systematic survey methodology}}  \cite{siddaway2019,watson2019}, with the aim to synthesise reproducibility guidelines from the existing AI literature into a comprehensive catalogue of recommendations that particularly foster reproducibility practices in modern machine learning.

There are several types of literature reviews that best suit different areas of science or engineering, and support different aims of the conducted review \cite{pare2015}.
A type of review that is focused on revealing  contradictions, inconsistencies, strengths and weaknesses on the given topic in the pre-existing literature is a \textbf{\textit{critical review}} \cite{snyder2019}.

Considering the multi-faceted nature of modern artificial intelligence, 
the choice of the critical review formula appears natural, even if we limit our survey to the area of machine learning.
However, as contemporary AI, and in particular,  machine learning, is characterised by a rapid pace of development, resulting in important findings and recent ideas being reported in non-archival sources, we include in our survey also some elements of a \textbf{\textit{narrative review}}  \cite{greenhalgh2005}. This allows us to summarise what has been written recently about various aspects of reproducibility, thus providing entry points for further individual studies by the readers.
In order to collect the literature for our survey, we have implemented the following steps.
\begin{itemize}
 \item {\bf Search for the literature.}
  We started with a systematic search for the relevant literature in the three most widely used bibliographical data bases:  Elsevier Scopus, Clarivate Web of Science and Google Scholar.
  Recall that Scopus and Web of Science collect mostly archival publications from journals, books and established conferences and have well-defined inclusion criteria for the papers that are covered.
  These databases focus on fundamental and natural sciences, including medicine, but their coverage of engineering disciplines is less comprehensive. This is particularly visible with respect to the recent areas of computer science, where many important conference and workshop publications are not included.
  Therefore we decided to search also the Google Scholar database, which has much more relaxed inclusion criteria.
  Google Scholar indexes recent and emergent sources, which are particularly relevant in the context of the aims of our survey. In addition, it also covers many archiving services collecting the so-called pre-prints or reports, such as e.g. arXiv, the use of which has recently become extremely popular among computer scientists.

 \item {\bf Inclusion criteria.}
  We only included papers written in English that concern machine learning reproducibility.
  Although we included literature from all disciplines, ranging from computer science and  information systems to medicine and natural sciences, we excluded studies on reproducibility for specific topics that did not focused on machine learning or artificial intelligence.
  Only papers published from 2017 to 2022 were included in the search, as we focused on the current state-of-the-art.
  Because each of the databases we considered has a different search mechanism, the queries were slightly different.
  For Scopus we started using the keywords \textit{\textbf{reproducibility AND (machine learning OR artificial intelligence)}}, searching only in the titles and abstracts.
  For each found paper, we initially determined the relevance by the title and abstract, excluding papers that  we deemed irrelevant, e.g. those from medicine which mentioned AI or ML but in fact focused on the  reproducibility of clinical results.
  After this initial pruning of Scopus results, a total of 35 papers were considered for the survey.  For Web of Science, we run the query \textit{\textbf{reproducibility AND artificial intelligence}}, matching the keywords in the title and the abstract of the papers published in the period 2017-2022. We obtained an initial set of 252 papers, whose list was exported and downloaded from the WoS system. Such a set was then checked by analysing the pertinence and relevance of each paper. Thirty-one papers out of the initial 252 turned out to be significant from this check. The excluded papers were either just brief abstract, tutorials or letters or were out of scope as dealing, for instance, with the application of AI in a specific domain (e.g. health and care) and the reproducibility was considered with respect to the generalizability of the domain results. For Google Scholar the search procedure was different  due to the more limited possibilities of asking complex queries with additional conditions and a different way of presenting the results of the search. We asked independently two queries \textit{\textbf{reproducibility of artificial intelligence}} and \textit{\textbf{reproducibility of machine learning}} and received very long lists of results: e.g. 87 600 answers to the first query. As these lists are presented in the relevance order proposed by Google Scholar, we analysed the first 10 web pages - so approx. 100 proposed papers. As a results we identified 39 potentially interesting paper links. After an expert inspection of their full files, we finally selected 28 papers for the deeper analysis.  
 
  Finally, after combining results from the three databases and excluding duplicated papers, we got a list of 81 relevant publications, among them 69 journal articles, with the rest  being conference papers, book chapters or arXiv pre-prints.

\item {\bf Final selection of papers.}
In order to decide which of these papers should be covered in the study
three of the four authors conducted independent evaluations of the 81 full-text papers,
agreeing to select 18 of them as highly relevant to the reproducibility in machine learning problem.
Any controversies while implementing this selection were discussed and resolved.
All these 18 papers are reviewed in our study, while some of the remaining 63
relevant sources found in the databases are cited as secondary examples,
particularly in the context of various applications of machine learning or artificial intelligence and the emerging reproducibility issues.
\end{itemize}

Besides the systematic search for relevant publications we used also a less formal method to get the papers for our survey, namely through backward and forward searches in the references of the papers we have identified in the first phase. 
Here the selection was less formal, as we were interested in highly influential papers on reproducibility with some relevance to machine learning or at least general artificial intelligence, in particular those papers that defined guidelines and recommendations about reproducibility practices. 
In particular, in this scenario we search more intensively the recent conference or journal guidelines or checkpoint lists (see section \ref{sec:guidelines}). 
Our aim in this phase of research was to investigate how and by whom the guidelines we can find in the current literature were proposed, and to find any recurring patterns, inconsistencies, or important gaps in those guidelines. 
Therefore, in this phase we didn't set a publication year census on the papers, including also older but seminal works.

\begin{figure}[htpb!]
    \centering
    \includegraphics[width=0.8\textwidth]{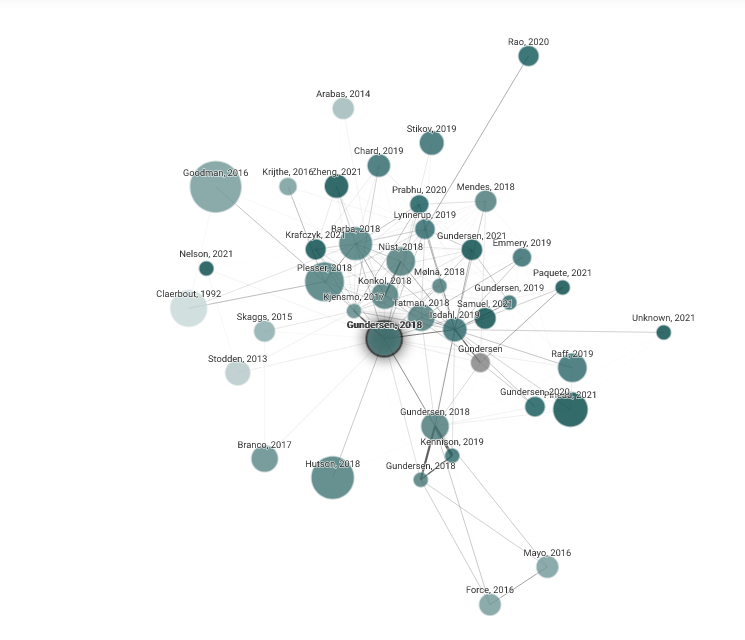}
    \caption{An example graph of connected papers concerning reproducibility,
     which was ``seeded'' by the work of Gundersen and Kjensmo \cite{gundersen_kjensmo_2018}.
     Papers that have highly overlapping references are considered more similar, 
     hence similar papers have strong connecting lines and cluster together. 
     Node size is the number of citations, while node colour is the publishing year (darker nodes are more recent)}
    \label{fig:citegraph}
\end{figure}

The search for most relevant citations in the analysed papers revealed relations between the existing works on reproducibility.
In order to further explore these relations and to identify the most influential papers and authors we explored a new tool to survey the literature and verify what we got.
This web-based visual tool, named Connected Papers \cite{connectedpapers}, allows the users to find papers relevant to a ``seed'' paper, which is entered as the search query.
Related papers are retrieved from the Semantic Scholar Paper Corpus
.
The similarity metric in Connected Papers is based on the concepts of co-citation and bibliographic coupling \cite{coupling2007}. 
Papers that have highly overlapping citations and references are considered as more similar, even if there are no direct citations between them.
The visualised graph clusters the found papers according to their similarity and highlights the shortest path from each node (paper) to the ``seed'' paper in similarity space.
Figure \ref{fig:citegraph} shows an example Connected Papers graph, which was built starting the search from one of the most relevant surveys about reproducibility in AI, the work of Gundersen and Kjensmo \cite{gundersen_kjensmo_2018}. 
In this case 40 papers were used by the program to build the graph, as those having the strongest connections to the origin paper.
When we inspected these papers, we have found that 17 were those already included in our references, while the remaining 23 were less relevant, covering reproducibility or replicability topics in areas other that ML and AI, or not meeting our inclusion criteria.
The graph reveals highly connected clusters of papers around the works co-authored by Gundersen, whom we identify as one of the most influential authors in the survey, and the works of Barba \cite{Barba}, Tatman et al. \cite{Tatman2018}, and Lynnerup et al. \cite{Lynnerup_et_al_2020}. 
Few papers, like the one by Konkol et al. \cite{konkol} are highly connected, but not relevant in our context.

\section{Terminology variety in biomedical applications}
\label{sec:term_biomedic}
In their work \cite{raw81}, Renard and co-authors referred to the definition by the National Academies of Science, Engineering, and Medicine \cite{NAS_2019}.
They focused explicitly on DL and the variability of DL model development, recommending researchers to detail the various sources on uncertainty, coming from the stochastic optimization, hyper-parameter setting, DL model architecture and the development middleware and/or infrastructure, as debated in the previous sections. 
In their work \cite{raw65}, Balagurunathan et al. focused mainly on repeatability (i.e., top quadrant of Figure \ref{fig:TerminologyDiagram}), as they claimed its importance as a first step to quantify the variability of an AI model and to gain confidence in the use of such a model. Nonetheless, they did not provide any recommendations or guidelines in this respect.

McDermott and co-authors introduced an additional specification of reproducibility, by differentiating among the following concepts \cite{raw57}:
\begin{itemize}
\item \emph{technical replicability}, which corresponds to the possibility to reproduce technically the AI-based method, yielding the same results reported in the paper describing the method. This can be mapped on the top-left quadrants of Figure \ref{fig:TerminologyDiagram} and entails aspects related to the sharing code and datasets.
\item \emph{statistical replicability}, which corresponds to the possibility to obtain the same results under re-sampled conditions that yield different technical configurations, but should not statistically affect the claimed result (e.g., a different set of random seeds, or train/test splits). This implication of reproducibility might be mapped onto the repeatability concept (i.e. top-left sub-quadrant), but adopting specific measures to cope with the variability and randomness of ML models. 
This is an issue of particular interest in the biomedical community as most of the scientific results are presented by considering their variance or confidence intervals, for instance, based on cross-validation strategies.  
\item \emph{conceptual replicability}, which corresponds to the possibility to reproduce the desired results under conditions that match the conceptual description of the original paper. This maps onto the two right quadrants of Figure \ref{fig:TerminologyDiagram} as it mainly corresponds to the \emph{generalizability} of the scientific results.
\end{itemize}

In his conference paper \cite{raw59}, Wojtusiak considered the reproducibility in a broad sense, ranging from replicability to robustness and generalizability (or corroboration), thus covering the four quadrants of Figure \ref{fig:TerminologyDiagram}. He provided valuable recommendations to foster the various dimensions of reproducibility and put emphasis on reporting the statistical variance of any outcomes (e.g., by reporting mean and variance of several runs or trials). This is because in biomedical applications the significance and acceptability of any results are usually considered from a statistical standpoint. Statistical analyses, actually, guide clinical researchers to draw reasonable and accurate inferences from the collected data and, thus, to make sound decisions in the presence of uncertainty, for instance with respect to the variability of response to treatment by diverse individuals. 

Similarly to McDermott et al., Wojtusiak debated the importance of generalizability of AI-based scientific results. Noteworthy, replicability, conceptual replicability, robustness or generalizability represent a very critical issue in biomedical applications, as without this capability any AI-based solutions may appear useless in clinical practice \cite{raw59}. The difficulties in ensuring this capacity come from peculiarities of the biomedical field, which is often referred to as the lack of robustness or reliability of the AI-powered tools, is due to the fact that AI models are often trained, tested and validated on hand-picked and limited datasets, often acquired by just one clinical institution and in a so-called \emph{retrospective} fashion (i.e., from the clinical records or repositories of the clinical institution, dating back even several years). These datasets are hence not enough representative of the data that can be encountered in clinical practice, as they can differ a lot from new data coming from other institutions or from those \emph{prospectively} acquired by the same institution (i.e., from the current clinical practice). These conditions are peculiar of the biomedical field and are  discussed in the main part of the paper.

\section{Reproducibility of ML/DL results in physical and simulated robotics research.}
\subsection{Reproducibility in robotics.}
In robotics, reproducible comparison to algorithmic baselines and previous results is particularly complicated because of the lack of reproducible experimental setups.
Whereas research in other areas of ML/DL, e.g. image processing, is usually benchmarked on large, \textit{\textbf{publicly available datasets}}, such as COCO\footnote{\url{https://cocodataset.org}} and The PASCAL Visual Object Classes\footnote{\url{http://host.robots.ox.ac.uk/pascal/VOC/}}, the datasets concerning robotics need to be more specialised, and they are often provided along with a particular research method or algorithm.
Such datasets, e.g. the KITTI Vision Benchmark Suite\footnote{\url{http://www.cvlibs.net/datasets/kitti/}} \cite{kitti2013}, commonly used in research related to self-driving cars and SLAM (Simultaneous Localisation and Mapping) offer pre-processed data, while the actual algorithms used for this pre-processing are often not specified in details (e.g. compensation of the vehicle motion while acquiring the 3D laser scanner data in KITTI).     
Another problem hampering reproducibility in robotics research using datasets can be exemplified by the case of KAIST Multi-Spectral Day/Night Data Set \cite{kaist2018dataset}, which is one of a very few datasets with visual spectrum and thermal images that can be applied for practical research in safety of autonomous cars and pedestrians. 
Hence, the KAIST dataset is widely used, but many works adopt modified variants, for example the “sanitized” version \cite{multispectral2018}, from which known annotation errors have been removed, or a subset of images with human silhouettes taller than 50 pixels.
As there is no apparent versioning of these modifications, research results using the KAIST dataset become hardly comparable in terms of quantitative metrics \cite{roszyk2022}.
\textit{These examples demonstrate that ensuring data provenance and traceability is also crucial in physical AI, while the metadata describing a given dataset should include a detailed description of the sensory data acquisition process and any post-processing that was applied afterwards}. 

In this review we can provide only few illustrative examples of problems specific to the reproduction of physical AI results, but even these examples show clearly that there is an urgent need of reproducible experimental setups and datasets, as well as performance metrics that are robust to the common factors of non-determinism that are present in this domain. 
Examples of \textit{\textbf{physical benchmarks}} designed to be reproducible can be found in robotic manipulation.
The ACRV Picking Benchmark \cite{leitner2017acrv} consists of a set of 42 common objects, a widely available shelf, and exact guidelines for object arrangement using stencils. 
An evaluation protocol is also provided in \cite{leitner2017acrv} in order to facilitate comparison of complete manipulation systems, including both perception and control aspects. 
The more recent Evolved Grasping Analysis Dataset (EGAD) \cite{morrison2020egad} serves the purpose of reproducible training and evaluation of robotic visual grasp algorithms.
A set of 49 diverse 3D-printable evaluation objects is provided to encourage physical, yet reproducible testing of robotic grasping algorithms. 

\subsection{Reproducibility of simulations.}
An alternative solution to the development of physical AI systems in real-world experiments is to use simulators \cite{trinkle2018}.
\textit{In contrary to the physical world, a simulated environment can be initialised in a deterministic way, and it is possible to reset the robot with it's control and learning algorithms to the default parameters and run a large number of independent experiments, which is essential in some domains, such as self-driving vehicles \cite{holen2022}}. 
However, these benefits come at the cost of possible overfitting of the trained ML/DL solution to the simulated environment. 
Models trained in simulation often transfer poorly to the physical world and fail on real robots, or their performance decreases substantially, which is known as the \textit{\textbf{reality gap}} problem, while the efforts towards transferring data-driven algorithms trained in simulations to the physical world are known as \textit{\textbf{sim2real}} \cite{sim2real,sim2realrss}.  
A common approach to bypass the reality gap is to \textit{\textbf{randomise}} these aspects of the feedback information that are challenging to simulate accurately.
Although this is an effective method to obtain robust learned models \textit{\textbf{transferable}} to real robots \cite{josifovski2022},
the use of extensive randomisation may compromise reproducibility.
Josifovski {\em et al.} \cite{josifovski2022} investigated and quantified the effects of randomisation on the learning of RL policies that transfer best to a real robot in a simple manipulation task, concluding that increasing the randomness of simulation parameters one can expect more robust models, but also a decrease in the ability of the algorithm to find a good policy in simulation.
Thus, less convergent learning experiments may bring different outcomes, becoming less reproducible. 

Simulations that are transferable to real robots, but still reproducible are researched in \cite{ferigo2020}.
This paper introduces Gym-Ignition, a framework to create reproducible robotic simulations for reinforcement learning experiments.  
The software architecture of Gym-Ignition avoids the use of \textit{\textbf{socket-based client-server communication}} between different modules,
as sockets can be preempted depending on the load of the operating system, which leads to non-determinism in the simulated environment feedback.
What is more, Gym-Ignition exposes to the users an interface to initialise all the random number generator seeds, which allows separate runs to be repeatable.
Ferigo {\em et al.} \cite{ferigo2020} compare the features of their simulator to a number of alternative solutions, finding also the recent Nvidia Isaac Gym\footnote{\url{https://developer.nvidia.com/isaac-gym}} \cite{issacgym2021} and Unity ML-Agents\footnote{\url{https://unity.com/products/machine-learning-agents}} as reproducible, while the OpenAI Gym \cite{openai2016} with OpenAI Robotic Environments\footnote{\url{https://openai.com/blog/ingredients-for-robotics-research/}}, which is commonly used to benchmark RL algorithms, does not fully support reproducibility. 

\end{screenonly}
\end{document}